\pdfoutput=1
\documentclass[11pt]{article}

\usepackage[final]{acl}
\usepackage{xcolor} 
\usepackage{times}
\usepackage{latexsym}
\usepackage{amsmath, amssymb, amsfonts}
\usepackage{caption} % 引入caption包
\usepackage[T1]{fontenc}
\usepackage{colortbl}            
\usepackage{booktabs}            
\usepackage{makecell}            
\usepackage{pifont}              
\usepackage{graphicx}            
\usepackage{caption}      
\usepackage{xcolor}
\usepackage[most]{tcolorbox}
\usepackage{algorithm}
\usepackage{algpseudocode}
\usepackage[utf8]{inputenc}
\usepackage{xcolor}
\usepackage{tcolorbox}
\usepackage{booktabs} % 确保在导言部分导入这个宏包

\usepackage{amssymb}
\usepackage{amsmath}
\definecolor{forestgreen}{RGB}{34,139,34}
\usepackage{microtype}

\usepackage{inconsolata}
\usepackage{balance}
\usepackage{graphicx}
\usepackage{multirow}
\definecolor{RowColor}{rgb}{0.97, 0.97, 1}
\usepackage{geometry}
\newcommand{\cmark}{\textcolor[rgb]{0.004, 0.663, 0}{\ding{51}}} 
\newcommand{\xmark}{\textcolor{red}{\ding{55}}}

\title{MDTeamGPT: A Self-Evolving LLM-based Multi-Agent Framework for Multi-Disciplinary Team Medical Consultation}

\author{
  Kai Chen$^1$, Xinfeng Li$^2$, Tianpei Yang$^1$$^*$, Hewei Wang$^3$, Wei Dong$^2$, Yang Gao$^1$ \\
  $^1$State Key Laboratory for Novel Software Technology, Nanjing University\\
  $^2$Nanyang Technological University \\
  $^3$School of Computer Science, Carnegie Mellon University \\
}

\begin{document}
\maketitle
\begin{abstract}
Large Language Models (LLMs) have made significant progress in various fields. However, challenges remain in  Multi-Disciplinary Team (MDT) medical consultations. Current research enhances reasoning through role assignment, task decomposition, and accumulation of medical experience. Multi-role collaboration in MDT consultations often results in excessively long dialogue histories. This increases the model’s cognitive burden and degrades both efficiency and accuracy. Some methods only store treatment histories. They do not extract effective experience or reflect on errors. This limits knowledge generalization and system evolution. We propose a multi-agent MDT medical consultation framework based on LLMs to address these issues. Our framework uses consensus aggregation and a residual discussion structure for multi-round consultations. It also employs a Correct Answer Knowledge Base (CorrectKB) and a Chain-of-Thought Knowledge Base (ChainKB) to accumulate consultation experience. These mechanisms enable the framework to evolve and continually improve diagnosis rationality and accuracy. Experimental results on the MedQA and PubMedQA datasets demonstrate that our framework achieves accuracies of 90.1\% and 83.9\%, respectively, and that the constructed knowledge bases generalize effectively across test sets from both datasets. Our code is publicly available at: \href{https://kaichennj.github.io/MDTeamGPT-Main}{https://kaichennj.github.io/MDTeamGPT}.

\end{abstract}

\section{Introduction}

When diagnosing and treating complex diseases, a single-specialty medical perspective frequently proves insufficient to address patients' comprehensive and precise healthcare needs, thereby necessitating the implementation of a Multi-Disciplinary Team (MDT) and diversified clinical perspectives~\cite{8,10}. MDT facilitates the development of more accurate and holistic therapeutic strategies through systematic integration of expertise across medical disciplines~\cite{7}. Nevertheless, the organization of interdisciplinary consultations demands substantial temporal and human resources, while remaining vulnerable to procedural oversights, cognitive biases, and communication inefficiencies~\cite{9}. Such systemic challenges may ultimately compromise healthcare quality and adversely affect patient outcomes.

\setlength{\textfloatsep}{10pt} % 控制图表与正文之间的垂直距离
\setlength{\floatsep}{10pt}     % 控制两个图表之间的垂直距离

\begin{figure}[t!]
  \captionsetup{skip=0pt} % 设置标题与正文的间距为0
  \centering
  \includegraphics[width=0.9\columnwidth]{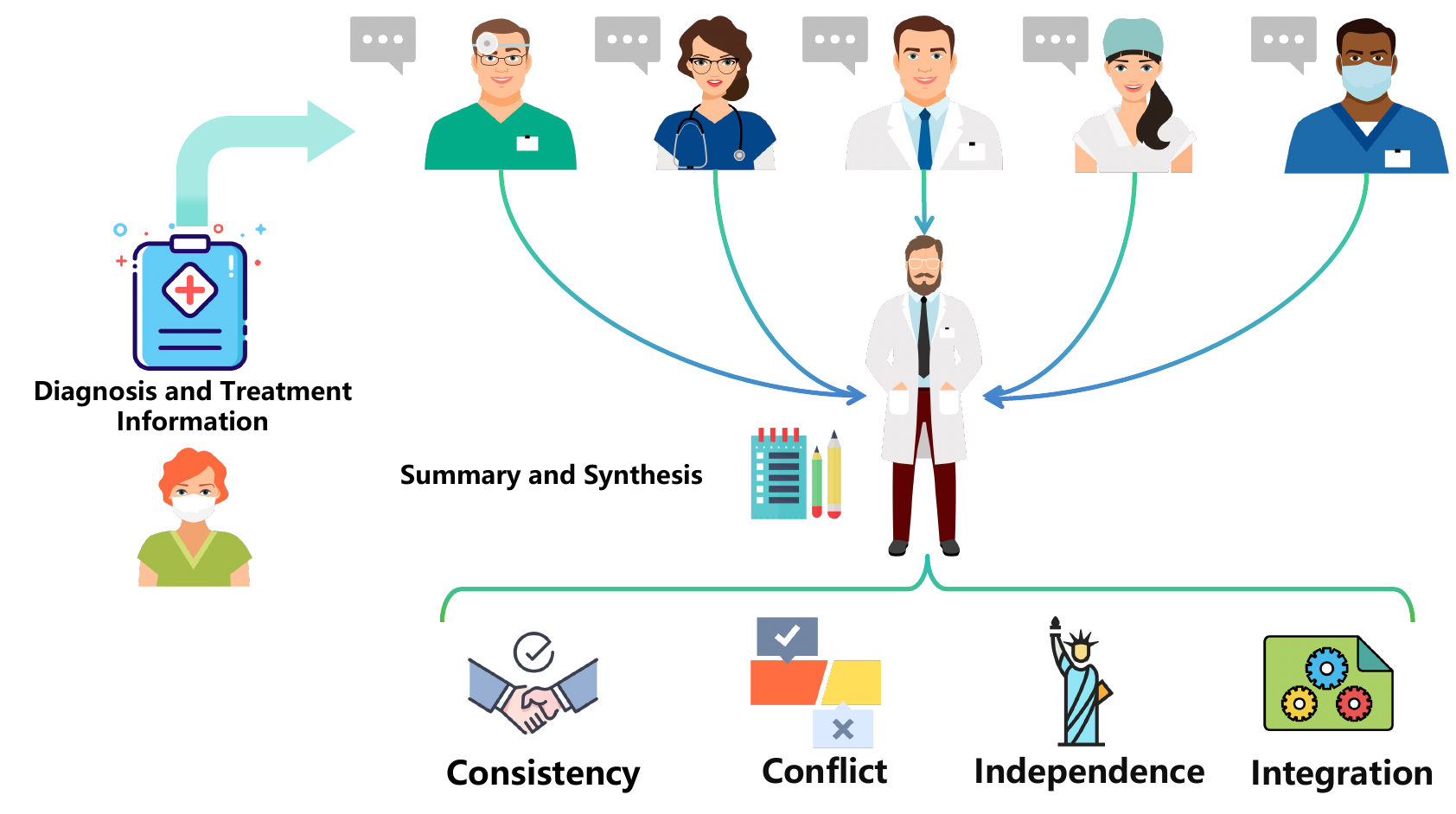}
  \caption{The lead physician consolidates and refines the discussion outcomes from the current round of agents by categorizing them into four detailed categories: consistency, conflict, independence, and integration, thereby enhancing the overall diagnostic clarity.}
  \label{fig1}
\end{figure}

Large Language Models (LLMs), with their massive parameter sizes and extensive training on vast knowledge bases from diverse domains, demonstrate strong reasoning abilities and zero-shot generalization capabilities~\cite{1,2}. These capabilities have led many researchers to explore the potential application of LLMs in medical MDT diagnostics~\cite{3,4}. Leveraging LLMs as autonomous agents offers a promising way to improve medical consultations by human doctors.

However, the application of LLMs in MDT diagnostics faces two primary challenges. First, privacy constraints and the lack of standardized data storage result in scarce medical consultation data, hindering the supervised fine-tuning (SFT) of large models. Secondly, although LLMs have been fine-tuned with medical knowledge, they remain prone to hallucination issues~\cite{5,6}.  In the medical field, such hallucinations can result in severe medical incidents even with minor errors~\cite{35}.

% The Multi-Agent technology based on LLMs effectively enhances the medical reasoning capabilities of LLMs \cite{11}. This design better induces the latent medical knowledge within LLMs, which cannot be solely elicited through the Chain-of-Thought (CoT) reasoning process. Additionally, through multi-round of interactions, Multi-Agent complements each other’s strengths, significantly reducing the likelihood of hallucination issues \cite{12}.

% Recent research on applying LLM-Multi-Agent in the medical domain has largely focused on role configuration exploration \cite{13, 30}, efficient subtask decomposition \cite{14}, the simulation of realistic clinical environments \cite{15}, and the accumulation of medical experience \cite{16,17}. Building on these foundational efforts, our work introduces a fresh perspective by examining two key areas. We first explore organizing and integrating observations from each LLM-Multi-Agent medical consultation to enhance reasoning, storing insights in a centralized historical pool. Second, we focus on leveraging correct and incorrect diagnostic experiences to improve the accuracy and adaptability of medical consultations.
Multi-agent based on LLMs enhances medical reasoning by stimulating latent medical knowledge through multi-round interactions and significantly reduces the occurrence of hallucinations~\cite{11,12}. Recent research in this domain has primarily focused on role configuration~\cite{13,30}, subtask decomposition~\cite{14}, clinical environment simulation~\cite{15}, and the accumulation of medical experience~\cite{16,17}. However, several challenges remain. First, multi-role interactions often result in excessively long dialogues. As contextual information accumulates, the cognitive load on the model increases, negatively impacting response efficiency and accuracy~\cite{40}. Second, some studies rely on simply storing diagnostic histories without effectively integrating correct diagnostic experiences or reflecting on errors, which may lead to biases when handling detailed information~\cite{41}. Finally, enhancing the generalization ability of systems in preserving medical experiences~\cite{26} is crucial for achieving self-evolution and leveraging collective intelligence.

In this work, we propose a self-evolving LLM-based multi-agent framework (MDTeamGPT) for multi-disciplinary team medical consultation. This framework incorporates various doctor roles and supporting roles, where a "patient" role arrives with specific background information and medical questions. The Primary Care Doctor assigns the patient to the most suitable specialist agents based on the patient’s condition. Subsequently, the specialist agents engage in a multi-round of discussion, sharing relevant information. After each round, a lead physician summarizes and organizes the discussion outcomes, categorizing them into \textbf{Consistency}, \textbf{Conflict}, \textbf{Independence}, and \textbf{Integration}, as illustrated in Figure \ref{fig1}. In the subsequent discussion rounds, agents refer to the information stored in the \textit{Historical Shared Pool} based on the residual discussion structure. Once a consensus is reached through multiple rounds of deliberation, the final consultation outcome is reviewed and filtered by a Safety and Ethics Reviewer before presenting the final results and recommendations. At this stage, the consultation outcomes are further categorized based on their correctness and stored in distinct knowledge bases, serving as references to enhance future consultations. Overall, the main contributions of our work are threefold:

\vspace{-0.3em}

\begin{itemize}
\item We propose an MDT-based multi-agent medical consultation framework that incorporates a lead physician mechanism and a residual discussion structure to effectively integrate multi-round discussion outcomes. This design reduces the cognitive load on agents, thereby enhancing the accuracy of reasoning and decision-making.
\vspace{-0.5em}
\item The framework employs an adaptive learning mechanism that leverages both correct and incorrect diagnostic experiences, progressively accumulating and strengthening transferable reasoning capabilities.
\vspace{-0.5em}
\item Experimental results demonstrate that the framework, utilizing 600 rounds of consultation experience, achieves accuracies of 90.1\% on the MedQA dataset and 83.9\% on the PubMedQA dataset. Cross-dataset tests further confirm that the knowledge bases retain generalizable reasoning skills, not just isolated consultation data.

\end{itemize}

\section{Related Work}

\subsection{LLM-based Multi-Agent Collaboration}

Well-structured multi-agent systems can significantly reduce errors and enhance interaction efficiency~\cite{18,19,20,21}. ChatDev~\cite{18} splits tasks into subtasks managed by an instructor and an assistant, using multi-round inquiries to mitigate hallucinations. MACNET~\cite{21} adopts a directed acyclic graph with topological sorting for interactive reasoning, and MetaGPT~\cite{23} encodes SOPs for roles akin to a software company, effectively coordinating specialized expertise. However, these methods primarily focus on software development and face applicability challenges in medical consultations. Medagents~\cite{13} assigns distinct doctor roles to each LLM-Agent and uses consensus voting, but lacks a robust strategy, risking collective hallucinations~\cite{22}. Additionally, these methods rely on static structures limited by zero-shot capabilities.

\subsection{Experience Utilization}
Inspired by human knowledge acquisition mechanisms, equipping LLM-Agents with reflective memory can enhance their problem-solving capabilities~\cite{24}. ExpeL~\cite{25} accumulates experiences from past successes and applies this knowledge during reasoning. Co-Learning~\cite{26} focuses on collecting experience-driven heuristics from prior actions, thus enabling agents to handle novel tasks more flexibly. IER~\cite{27} permits LLM-Agents to iteratively refine accumulated experiences throughout task execution. Selfevolve~\cite{28} employs LLMs both as knowledge providers and as self-reflective programmers, allowing agents to evolve through reflective processes. Agent Hospital~\cite{29} utilizes medical record libraries and experience databases to accumulate diagnostic data, enhancing prompts for medical agents and supporting their evolution. However, the lack of abstraction, summarization, and reflection on erroneous cases limits the ability to fully leverage valuable insights from mistakes.

\subsection{Discussion and Aggregation}
An effective discussion process can significantly enhance LLM-Multi-Agent performance. The Nominal Group Technique (NGT)~\cite{31} is a structured decision-making method that moves teams from individual ideation to collective consensus through four stages: personal idea generation, round-robin sharing, focused discussion, and voting. Building on this principle, LLM Discussion~\cite{32} augments the creative potential of LLMs by dividing conversations into initiation, discussion, and convergence phases, thereby promoting iterative and divergent exchanges among multiple agents.

% The first line of the file must be
% \begin{quote}
% \begin{verbatim}
% \documentclass[11pt]{article}
% \end{verbatim}
% \end{quote}

% To load the style file in the review version:
% \begin{quote}
% \begin{verbatim}
% \usepackage[review]{acl}
% \end{verbatim}
% \end{quote}
% For the final version, omit the \verb|review| option:
% \begin{quote}
% \begin{verbatim}
% \usepackage{acl}
% \end{verbatim}
% \end{quote}

% To use Times Roman, put the following in the preamble:
% \begin{quote}
% \begin{verbatim}
% \usepackage{times}
% \end{verbatim}
% \end{quote}
% (Alternatives like txfonts or newtx are also acceptable.)

% Please see the \LaTeX{} source of this document for comments on other packages that may be useful.

% Set the title and author using \verb|\title| and \verb|\author|. Within the author list, format multiple authors using \verb|\and| and \verb|\And| and \verb|\AND|; please see the \LaTeX{} source for examples.

% By default, the box containing the title and author names is set to the minimum of 5 cm. If you need more space, include the following in the preamble:
% \begin{quote}
% \begin{verbatim}
% \setlength\titlebox{<dim>}
% \end{verbatim}
% \end{quote}
% where \verb|<dim>| is replaced with a length. Do not set this length smaller than 5 cm.

\section{Method}

\setlength{\textfloatsep}{5pt} % 控制图表与正文之间的垂直距离
\setlength{\floatsep}{5pt}     % 控制两个图表之间的垂直距离

\begin{figure*}[t]
 \captionsetup{skip=0pt} % 设置标题与正文的间距为0
\centerline{\includegraphics[width=0.95\textwidth]{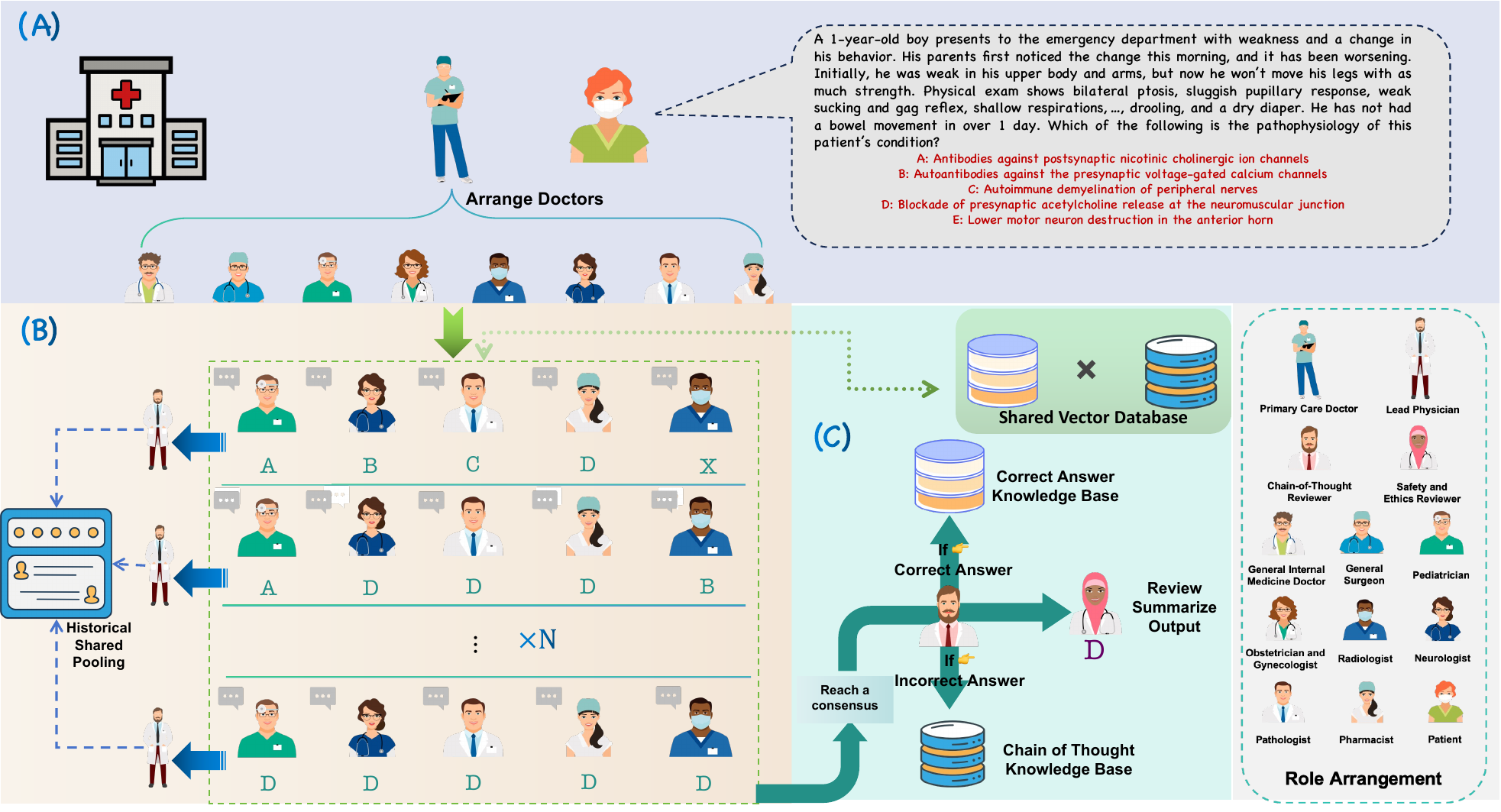}}
  \caption {Overview of our MDTeamGPT medical consultations framework:
(A) arranging strategically specialist doctors based on the patient’s specific condition;
(B) orchestrating multi-round 
 collaborative consultations with organized clinical information;
(C) summarizing and outputting the final diagnostic and treatment recommendations.}
\label{fig2}
\end{figure*}

In this section, we provide a detailed description of our medical consultation framework. Within this framework, we have established a set of specialist doctor Agent roles,   \( \mathcal{S} \) (General Internal Medicine Doctor, General Surgeon, Pediatrician, Obstetrician and Gynecologist, Radiologist, Neurologist, Pathologist, Pharmacist), and a set of auxiliary Agent roles,  \(\mathcal{A}\)  (Primary Care Doctor, Lead Physician, Chain-of-Thought Reviewer, Safety and Ethics Reviewer). The roles within  \(\mathcal{A}\)  do not directly participate in the consultation process; their functions are elaborated in Sections 3.2 and 3.3. This role setup broadens the framework’s applicability to diverse complex medical cases. The prompt templates for all roles are provided in Appendix \ref{Prompt Templates}. Figure \ref{fig2} presents an overview of the proposed framework, which is divided into three critical stages.

\subsection{Aranging Specialist Doctors}

When the patient agent comes for a consultation carrying personal background \( C \) and medical problem \( Q \), the Primary Care Doctor Agent assigns specialist doctor Agents based on the specific circumstances of the patient. Before making the final decision, the Primary Care Doctor Agent provides \textit{Reasons} for the selection. To ensure the triage doctor's output is more accurate and structured, we configure the Primary Care Doctor Agent with a few-shot example as a reference. The workflow is as follows:
This description can be expressed by the following formula:

\vspace{-1.5em}

\begin{equation}
\text{Reasons}, \text{Roles} = \text{LLM}\left(\text{Agents} \mid C, Q\right)
.\label{eq1}\end{equation}
\begin{equation}
\text{Roles} \subseteq \mathcal{S}
.\label{eq2}\end{equation}

This approach prevents information contamination caused by an excess of irrelevant expert agents. During configuration, we provide the Primary Care Doctor Agent with a limited number of cases, encouraging it to output responses in a prescribed format. An example is presented below:
\begin{center}
\begin{tcolorbox}[
    colback=forestgreen!10!white,  % 背景色：淡森绿色
    colframe=forestgreen!80!black, % 边框色：深森绿色
    boxrule=0.5mm, 
    boxsep=2mm, 
    left=2mm, 
    right=2mm, 
    top=2mm, 
    bottom=2mm, 
    width=\columnwidth,
    title=\small Primary Care Doctor Agent Output Example Template % 添加标题
]
\scriptsize
\textbf{Input question:} A 1-year-old boy presents to the emergency department with weakness and a change in his behavior. His parents first noticed the change this morning, and it has been worsening. Initially, he was weak in his upper body and arms, but now he won’t move his legs with as much strength. Physical exam shows bilateral ptosis, sluggish pupillary response, weak sucking and gag reflex, shallow respirations,…, drooling, and a dry diaper. He has not had a bowel movement in over 1 day. Which of the following is the pathophysiology of this patient’s condition?

\vspace{0.5em}

\textbf{Primary Care Doctor :}

• <\texttt{Reasons}>

• <\{\texttt{General Internal Medicine Doctor}\}, \{\texttt{Obstetrician and Gynecologist}\}, \{\texttt{Radiologist}\}, \{\texttt{Pathologist}\}, \{\texttt{Pharmacist}\}>

\end{tcolorbox}
\end{center}

\subsection{Multi-Round Consultations}

Once the specialist doctors for the consultation have been determined, the consultation process officially begins. In the initial round, each specialist doctor presents their opinion based on the patient's condition, denoted as \( S_{1,k} \) (where \( S_{1,k} \) represents the response of the \( k \)-th specialist doctor in the first round). Each specialist doctor also provides an option ID along with the corresponding content for the issue at hand. At this stage, each specialist doctor agent cannot observe the statements made by other agents.

\begin{algorithm}
\caption{Multi-Round Medical Consultation}
\footnotesize % 统一缩小整个算法字体
\begin{algorithmic}[1] % The number [1] ensures that lines are numbered

\State \textbf{Initialize:} Speech pool \( S_1 = \{s_{1,1}, s_{1,2}, \dots, s_{1,n}\} \)
\vspace{0.5em}
\State \textbf{Round 1:}
\For{each specialist \( k \)}
    \State \( S_{1,k} \gets f(C, Q) \) \Comment{Each specialist provides an initial response}
\EndFor
\State \( S_1^4 \gets \text{LeadPhysician}(S_1) \) \Comment{Process into four viewpoints}
\State Store \( S_1^4 \) in \( H \) 
\vspace{0.5em}
\State \textbf{Round 2:}
\For{each specialist \( k \)}
    \State \( S_{2,k} \gets f(S_1^4, C, Q) \)
\EndFor
\State \( S_2^4 \gets \text{LeadPhysician}(S_2) \)
\State Store \( S_2^4 \) in \( H \)
\vspace{0.5em}
\State \textbf{Subsequent Rounds:}
\State Set \( i = 2 \)

\While{not \( \text{Consensus}(S_n) \) \textbf{and} \( i \leq \text{MaxRounds} \)}
    \For{each specialist \( k \)}
        \State \( S_{i+1,k} \gets g(S_i^4, S_{i-1}^4, C, Q) \) \Comment{Use processed viewpoints from last two rounds}
    \EndFor
    \State \( S_{i+1}^4 \gets \text{LeadPhysician}(S_{i+1}) \) 
    \State Store \( S_{i+1}^4 \) in \( H \)
    \State Increment \( i \)
    
    \State \textbf{Consensus Check:}
    \If{\( \forall k, m \in \{1, \dots, n\} : S_{i,k} = S_{i,m} \)}
        \State Set \( \text{Consensus}(S_i) = \text{True} \)
    \EndIf
\EndWhile
\vspace{0.5em}
\If{not \( \text{Consensus}(S_i) \)}
    \State \textbf{Decision by Majority Rule:}
    \If{tie in votes}
        \State Randomly select one answer from agents' choices
    \EndIf
\EndIf

\end{algorithmic}
\end{algorithm}

All specialists' statements are sent to the lead physician(as detailed in Appendix \ref{Lead Physician Agent’s Work}), who processes them into four distinct components: \textit{Consistency}, \textit{Conflict}, \textit{Independence}, and \textit{Integration}.

\begin{itemize}
    \item \textbf{Consistency}: Aggregates the parts of individual statements that are consistent across multiple agent statements.
    \vspace{-0.5em}
    \item \textbf{Conflict}: Identifies and lists conflicting points between different statements when the answers are not uniform; otherwise, this component remains empty.
    \vspace{-0.5em}
    \item \textbf{Independence}: Extracts the unique viewpoints of each agent, i.e., aspects mentioned by a specific agent but not by others; Remains empty if no unique views exist.
    \vspace{-0.5em}
    \item \textbf{Integration}: Synthesizes all agent statements into a cohesive, well-structured summary incorporating all perspectives.
\end{itemize}

The processed statements are denoted as \( S_1^4 \) and are subsequently stored in the \textit{Historical Shared Pool} (\( H \)). Once statements are stored, the consultation moves to the next round.

Starting from the second round, each specialist doctor agent can access the structured statements from the previous round, stored in the \(H\). They integrate these insights to enhance their prompts and formulate their responses, denoted as \( S_{2,k} \), while selecting an option ID and corresponding answer content for the given issue. As in the first round, all agents' statements are sent to the lead physician, who processes them into \( S_2^4 \).

% \vspace{-0.14em}

From round \( i+1 \) (where \( i \geq 2 \)), specialist doctor agents can review the structured statements from rounds \( i \) and \( i-1 \), denoted as \( S_i^4 \) and \( S_{i-1}^4 \), respectively. By integrating insights from the previous two rounds, they refine their prompts, articulate their viewpoints, and provide an option ID and corresponding content for further discussion. The discussion continues until all specialist doctor agents reach a consensus on the final answer. If a consensus is not achieved and the number of discussion rounds has not reached the predefined upper limit (set at 10 rounds), the discussion proceeds to the next round. If the maximum number of rounds is reached without achieving consensus, the final decision is made based on the \textit{majority rule}. If votes are evenly distributed, the final answer is randomly selected from the agents' choices.

\setlength{\textfloatsep}{2pt} % 控制图表与正文之间的垂直距离
\setlength{\floatsep}{2pt}     % 控制两个图表之间的垂直距离
\begin{figure}[!htbp]
  \captionsetup{skip=-3pt} % 设置标题与正文的间距为0
  \centering
  \includegraphics[width=0.4\columnwidth]{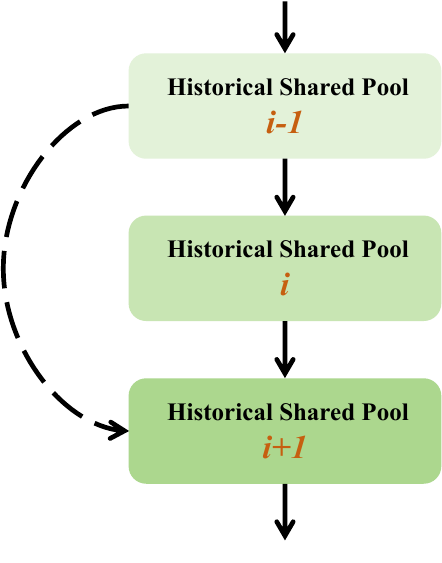}
  \caption{Diagram of the residual discussion structure in the MDT medical consultation process.}
  \label{HSP}
\end{figure}
\vspace{-0.5em}
This \textbf{consensus aggregation}, combined with the \textbf{residual discussion structure} (see Figure \ref{HSP}), reduces information pollution, enhances discussion efficiency, and minimizes the cognitive load on the model. The structure limits specialist doctor agents’ access to deeper memory layers, helping to prevent excessive influence from other agents and thereby mitigating hallucinations to some extent. The consultation process follows Algorithm 1.

\subsection{Summary and Output Stage}

At this stage, the final output (\( C \)) is reviewed by the \textit{Safety and Ethics Reviewer} agent. This agent refines and verifies the consultation outcome, identifies any potential safety concerns, and ultimately determines the final conclusion (\( R \)). The conclusion is later validated against the correct result. 

If the consultation outcome is accurate, the \textit{Chain-of-Thought Reviewer} summarizes and organizes the patient's background information(\( B \)), medical problem(\( Q \)), and the structured statements from the final consultation round \( S_{\text{final}}^4 \). This processed information is then archived into the \textit{Correct Answer Knowledge Base (\textit{CorrectKB})}. The storage format is as follows:

\begin{center}
\begin{tcolorbox}[
    colback=forestgreen!10!white,  % 背景色：淡森绿色
    colframe=forestgreen!80!black, % 边框色：深森绿色
    boxrule=0.5mm, 
    boxsep=2mm, 
    left=2mm, 
    right=2mm, 
    top=2mm, 
    bottom=2mm, 
    width=\columnwidth,
    title=\small CoT Reviewer Storage Format (Correct Answer) % 设置标题并调整大小
]
\vspace{-0.5em}

\small % 统一字体大小
\texttt{Chain-of-Thought Reviewer(\{C, B, Q, S\textsubscript{final}\textsuperscript{4}\}):}

\vspace{0.5em}

\texttt{
\{"Question": <...>,\\
 "Answer": <...>,\\
 "Summary of S\textsubscript{final}\textsuperscript{4}": <...>\}
}

\end{tcolorbox}
\end{center}

Conversely, if the consultation outcome is incorrect, the \textit{Chain-of-Thought Reviewer} generates an abstract summary of the session. This summary includes \( B \) and \( Q \) and is structured according to the initial hypothesis, analysis process, final conclusion, and reflection on error causes. The summary is then stored in the \textit{Chain-of-Thought Knowledge Base (\textit{ChainKB})}. The storage format is as follows:

\begin{center}
\begin{tcolorbox}[
    colback=forestgreen!10!white,  % 背景色：淡森绿色
    colframe=forestgreen!80!black, % 边框色：深森绿色
    boxrule=0.5mm, 
    boxsep=2mm, 
    left=2mm, 
    right=2mm, 
    top=2mm, 
    bottom=2mm, 
    width=\columnwidth,
    title=\small CoT Reviewer Storage Format(Incorrect Answer) % 设置标题并调整大小
]
\vspace{-0.5em}

\small % 统一字体大小
\texttt{Chain-of-Thought Reviewer(\{C, B, Q, S\textsubscript{i}\textsuperscript{4}, ..., S\textsubscript{final}\textsuperscript{4}\}):}

\vspace{0.5em}

\texttt{
\{"Question": <...>, \\
  "Correct Answer": <...>,\\
 "Initial Hypothesis": <...>,\\
 "Analysis Process": <...>,\\
 "Final Conclusion": <...>,\\
 "Error Reflection": <...>\}
}

\end{tcolorbox}
\end{center}

% When a new patient arrives, their \( B \) and \( Q \) are used to retrieve the most similar cases from both knowledge bases based on \textit{cosine similarity computation}. This retrieval process optimizes the prompt (\( P \)) for the specialist doctor agents.  

When a new patient arrives, their \( B \) and \( Q \) are used to retrieve the \(top-K\)  most similar cases from both knowledge bases based on \textit{cosine similarity computation} (\(K\) = 5  in our application). This retrieval process enhances the prompts (\( P \)) for the specialist doctor agents.

To maintain the independent reasoning of each specialist doctor agent, these two knowledge bases are not referenced during the initial discussion round. Instead, they are utilized starting from the second round, particularly in cases where conflicting opinions arise. However, if a consensus is reached in the first round, these knowledge bases may be referenced as a reflective measure after the discussion concludes. The full process is detailed in Algorithm 2.

\begin{algorithm}
\caption{Summary and Evolution}
\small
\begin{algorithmic}[1] % Lines are numbered

\State \textbf{Input:} \( C \), \( B \), \( Q \), \(\text{\textit{CorrectKB}}\), \(\text{\textit{ChainKB}}\)

\vspace{0.5em}
\State \textbf{Outcome Validation:}
\If{\(\text{Valid}(R)\)}
    \State \( \text{Record} \gets \text{CoT\_Reviewer}(\{C, B, Q, S_{\text{final}}^4\}) \)
    \State \(\text{Store}(\text{Record}, \text{\textit{CorrectKB}})\) 
\Else
    \State \( \text{Record} \gets \text{CoT\_Reviewer}(\{C, B, Q, S_{i}^4, ..., S_{\text{final}}^4\}) \)
    \State \(\text{Store}(\text{Record}, \text{\textit{ChainKB}})\) 
\EndIf
\vspace{0.5em}
\State \textbf{Prompt Enhancement:}
\State $similarity \gets \text{CosineSim}(B, \text{\textit{CorrectKB}}, \text{\textit{ChainKB}})$
\State $TopK\_similarity \gets \text{SelectTopK}(similarity, K)$
\State $P \gets \text{Retrieve}(TopK\_similarity, \text{\textit{CorrectKB}}, \text{\textit{ChainKB}})$
\State $enhancedPrompt \gets \text{Enhance}(P)$
\vspace{0.5em}
\State \textbf{Application of Enhanced Prompt:}
\State \( \text{round} \gets 1 \) 
\While{\textbf{not} \( \text{Consensus}(C) \)}
    \State \( \text{UsePrompt}(enhancedPrompt, \text{round}) \)
    \State \( \text{round} \gets \text{round} + 1 \)
    \If{\(\text{round} = 2\) \textbf{and} \text{conflicts arise}}
        \State \( \text{ConsultKBs}(P) \)
    \EndIf
\EndWhile
\vspace{0.5em}
\If{\(\text{round} = 1\) \textbf{and} \(\text{Consensus}(C)\)}
     \State \( \text{ConsultKBs}(P)\) 
\EndIf

\end{algorithmic}
\end{algorithm}

It is worth noting that the core model responsible for embedding textual content into the vector spaces of \textit{CorrectKB} and \textit{ChainKB} is OpenAI’s `text-embedding-3-small'\footnote{\url{https://platform.openai.com/docs/guides/embeddings/embedding-models}}.

\section{Experiments}
% \label{sec:bibtex}
\subsection{Datasets}
We utilize the MedQA~\cite{33} and PubMedQA~\cite{34} datasets to validate our framework. The MedQA dataset consists of USMLE-style questions, each offering four or five possible answers, designed to assess medical knowledge and practical skills. PubMedQA, based on research paper abstracts, presents questions with Yes/No/Maybe answers, aiming to evaluate the performance of natural language processing models in academic question answering. The final results are based on the test sets of each dataset, with the \textit{CorrectKB} and \textit{ChainKB} containing only experiences from the training sets.

\subsection{Main Results}

In this experimental section, we primarily investigate the zero-shot accuracy and F1-score advantages of our proposed MDTeamGPT framework in the context of medical consultations (additional metrics were assessed through human evaluation, as detailed in Appendix~\ref{Human evaluation}). Under this framework, 600 consultation rounds were conducted, during which the experiences were stored in \textit{CorrectKB} and \textit{ChainKB} (MedQA and PubMedQA each utilized their respective training sets for these consultations). This benchmark was chosen after considering both performance and cost. The overall results are shown in Table \ref{table1}, where gpt-4-turbo serves as the base model across all configurations.

\textbf{Single-Agent} denotes the baseline performance using only gpt-4-turbo; \textbf{Single-Agent (w/) CoT} incorporates a "Let’s think step by step" approach in the answering process; \textbf{Single-Agent (w/) ReAct} employs a ReAct-style prompt, simplified and modified in our experiments to consist of the "Thought", "Action (Selection)", and "Observation (Reflection)" processes. Although our approach achieves slightly lower accuracy on the MedQA dataset than Medprompt~\cite{35}—likely because Medprompt is evaluated solely on four-option scenarios—it outperforms Medprompt on the PubMedQA dataset and in overall averages. Overall, MDTeamGPT demonstrates strong average accuracy, proving its effectiveness for advanced LLM-based medical applications.

\begin{table*}[!htbp]
\centering
\makebox[0.8\textwidth][c]{%
\scalebox{0.8}{%
\begin{tabular}{l|cccc|c}
\specialrule{1.2pt}{0pt}{0pt} % 上面加粗
\multirow{2}{*}{\textbf{Method}}                                          & \multicolumn{2}{c}{MedQA}     & \multicolumn{2}{c|}{PubMedQA} & Average       \\ \cline{2-5}
                                                 & Accuracy (\%)  & F1-score (\%)  & Accuracy (\%)  & F1-score (\%)  & Accuracy (\%)  \\ \hline
Single-Agent                                     & 77.4          & 76.2          & 75.3          & 73.7          & 76.4          \\
Single-Agent (w/) CoT                            & 76.6          & 75.1          & 76.9          & 74.8          & 76.8          \\
Single-Agent (w/) ReAct                          & 78.5          & 76.9          & 75.8          & 74.1          & 77.2          \\
Medprompt~\cite{35}                                & \textbf{90.2} & \textbf{88.9}          & 82.0          & 80.3          & \underline{86.1}          \\ 
Multi-expert Prompting~\cite{30}                  & 86.6          & 84.9          & \underline{82.7}          & \underline{81.6}          & 84.2          \\ \hline
LLM Discussion~\cite{32}                          & 80.4          & 78.3          & 78.2          & 76.4          & 79.3          \\
MedAgents~\cite{13}                               & 83.7          & 82.1          & 76.8          & 75.1          & 80.3          \\ \hline
\colorbox{green!8}{MDTeamGPT (\textbf{ours})}                    & \underline{90.1}          & \underline{88.4} & \textbf{83.9} & \textbf{82.6} & \textbf{87.0} \\ 
\specialrule{1.2pt}{0pt}{0pt} % 下面加粗
\end{tabular}%
}%
}
\caption{\label{table1} Main results on accuracy and F1-score across MedQA and PubMedQA datasets (all results were obtained using gpt-4-turbo). We highlight the optimal and suboptimal methods in \textbf{bold} and \underline{underline}, respectively.}
\end{table*}

\begin{table*}[!htbp]
\centering
% 设置从第一行开始，每隔一行着色为 RowColor，其他为白色
\rowcolors{2}{RowColor}{white}
\makebox[\textwidth][c]{% 调整宽度至全文宽度
\scalebox{0.89}{% 缩放因子
\begin{tabular}{cccc|cc|c}
\specialrule{1.2pt}{0pt}{0pt} % 加粗顶线
\makecell{Residual \\ discussion mode} & \makecell{Lead \\ Physician} & \makecell{\textit{CorrectKB} \\ (600 rounds)} & \makecell{\textit{ChainKB} \\ (600 rounds)} & MedQA (\%) & PubMedQA (\%) & Average (\%) \\ \hline
\xmark  & \xmark  & \xmark  & \xmark  & 78.5      & 74.0        & 76.3        \\
\cmark & \xmark  & \xmark  & \xmark  & 76.9      & 74.8        & 75.9        \\
\xmark  & \cmark & \xmark  & \xmark  & 78.2      & 75.4        & 76.8        \\
\cmark & \cmark & \xmark  & \xmark  & 84.3      & 77.6        & 81.0        \\
\cmark & \cmark & \cmark & \xmark  & \underline{87.3}      & \underline{83.5}        & \underline{85.4}        \\
\cmark & \cmark & \xmark  & \cmark & 85.2      & 78.1        & 81.6        \\
\cmark & \cmark & \cmark & \cmark & \textbf{90.1}      & \textbf{83.9}        & \textbf{87.0}        \\ \hline
\specialrule{1.2pt}{0pt}{0pt} % 加粗底线
\end{tabular}%
}%
}
\caption{\label{table2}Ablation study evaluating the impact of different module compositions on accuracy metrics.}
\end{table*}

\subsection{Ablation Studies}

To examine the contribution of each component within the MDTeamGPT framework, we conducted experiments combining different configuration settings. Throughout these experiments, gpt-4-turbo served as the backbone model for all agents.

In Table \ref{table2}, the first experimental setting permits all agents to engage in a joint discussion without any specific discussion strategy; the system only terminates when either a consensus is reached among all agents or the maximum number of discussion rounds is exhausted. The second experiment introduces the residual discussion mode, resulting in lower accuracy on MedQA but a 0.8\% improvement on PubMedQA, leading to a 0.4\% decrease in average accuracy compared with the first experiment. In the third experiment, we remove the residual discussion mode and only employ the lead physician to synthesize discussion results. Relative to the first experiment, this yields a 0.3\% drop on MedQA but a 1.4\% increase on PubMedQA, translating into a 0.5\% rise in average accuracy.

Notably, combining the residual discussion mode and the lead physician in the fourth experiment raises the accuracy on MedQA and PubMedQA by 5.8\% and 3.6\%, respectively, compared with the first experiment, demonstrating that jointly applying residual discussion mode and lead physician offers substantial enhancements in consultation efficiency and accuracy.

We further investigate the role of the two knowledge bases (\textit{CorrectKB} and \textit{ChainKB}) in Experiments 5, 6, and 7, where both knowledge bases store experiences from 600 consultation rounds. Using only \textit{CorrectKB} yields 2.1\% and 5.4\% higher accuracy on MedQA and PubMedQA, respectively, compared to using only \textit{ChainKB}, and leads to a 3.8\% increase in overall accuracy. This finding aligns with intuition, as \textit{CorrectKB} Retain verified correct consultation experiences that can better guide subsequent diagnoses. However, the combined use of both \textit{CorrectKB} and \textit{ChainKB} achieves an accuracy superior to either \textit{CorrectKB} or \textit{ChainKB} alone, corroborating our hypothesis that integrating correct experiences with reflective insights maximally boosts consultation accuracy.

\subsection{Self-Evolving}

\begin{figure*}[t]
\centerline{\includegraphics[width=0.8\textwidth]{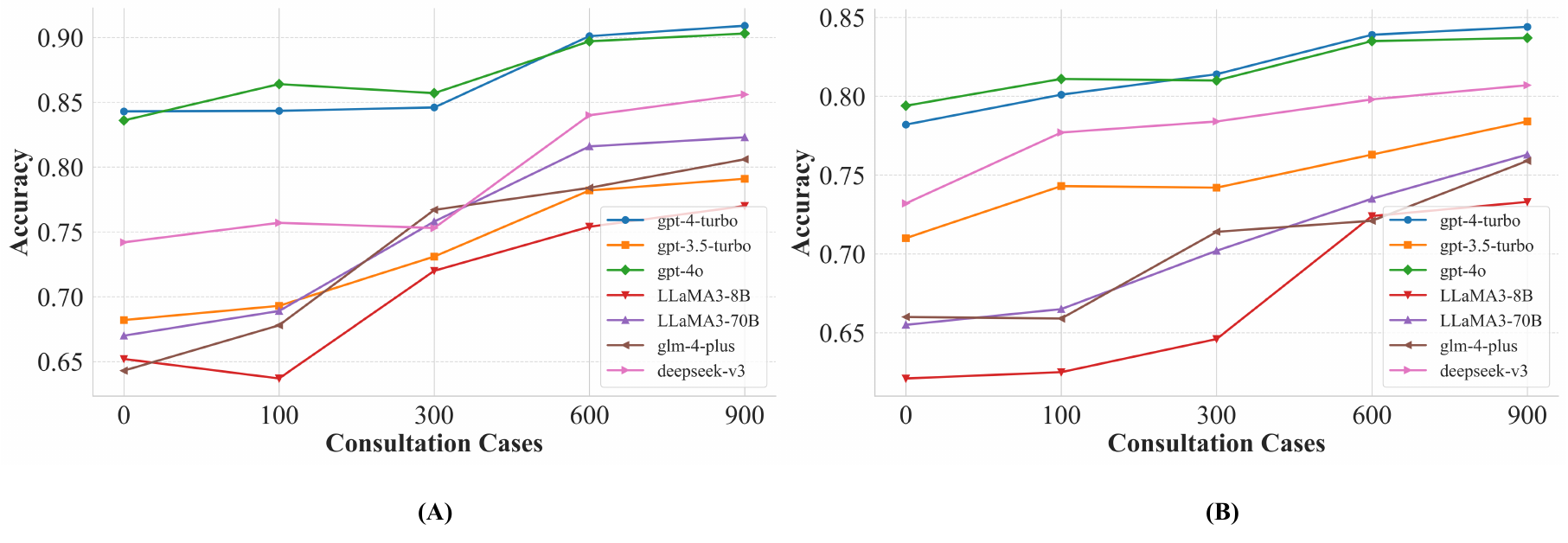}}
  \captionsetup{skip=0pt} % 设置标题与正文的间距为0
  \caption {Demonstration of the self-evolving capability of our proposed MDTeamGPT framework. As the number of consultation rounds increases, the framework progressively refines its diagnostic accuracy. (A) presents the performance of our MDTeamGPT tested on MedQA, and (B) illustrates corresponding results on PubMedQA.}
\label{fig3}
\end{figure*}

\begin{figure*}[!htbp]
\centerline{\includegraphics[width=0.8\textwidth]{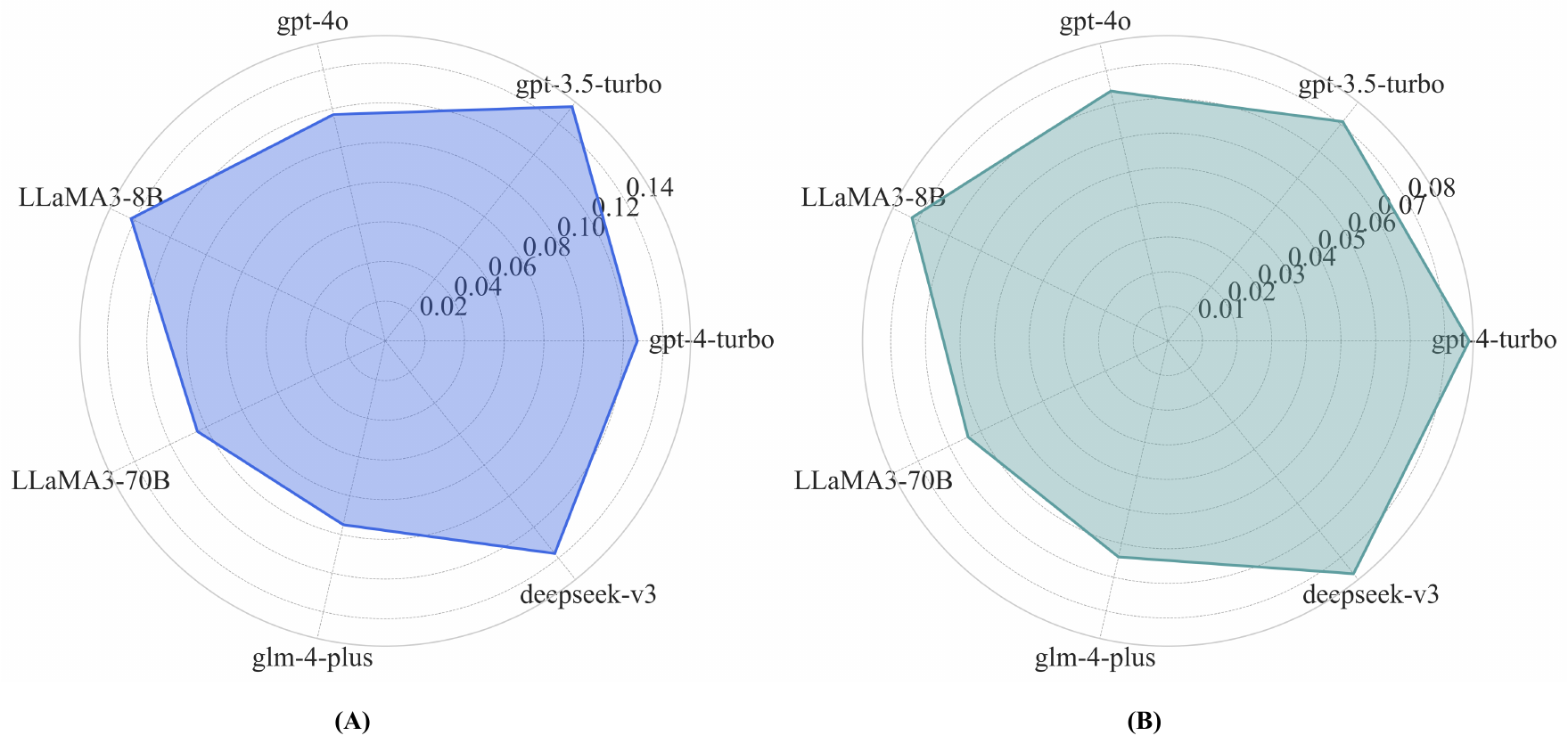}}
 \captionsetup{skip=0pt} % 设置标题与正文的间距为0
  \caption {Accuracy enhancements achieved by MDTeamGPT over single-agent baselines across multiple LLMs. (A) depicts the performance boost on MedQA, and (B) shows analogous gains on PubMedQA. These findings highlight the framework’s adaptability across language models and robust efficacy in medical consultation scenarios.}
\label{fig4}
\end{figure*}

In this section, we evaluate several LLMs — gpt-4-turbo, gpt-3.5-turbo, gpt-4o, LLaMA3-8B, LLaMA3-70B~\cite{36}, glm-4-plus~\cite{37}, and deepseek-v3~\cite{38}—as the core models for our MDTeamGPT framework, testing their self-evolving capability. Figure \ref{fig3} (A) and (B) show how the test accuracy on MedQA and PubMedQA changes with the increase in consultation samples and stored case experiences. Regardless of the LLMs used, accuracy improves with more consultation samples and stabilizes after approximately 600 cases. Figure \ref{fig3} shows an interesting observation: LLMs with relatively weaker foundational abilities, such as LLaMA3-8B, improve faster (steeper slope) with accumulated consultation experience compared to stronger models like gpt-4-turbo and gpt-4o.

\subsection{Framework Efficacy}
% To evaluate the effectiveness of MDTeamGPT across different LLMs, we experimented to assess the framework’s performance. The knowledge bases used in this experiment stored 600 rounds of medical consultation experiences. Figure \ref{fig4} (A) and (B) compare the accuracy improvements of MDTeamGPT versus single-agent baselines for each model. The results show that all models experience varying degrees of improvement after using MDTeamGPT, highlighting the framework's broad applicability and effectiveness. Although the extent of improvement varies with the model’s static knowledge and instruction-following capabilities, MDTeamGPT significantly enhances medical consultation across different LLMs.

To evaluate the effectiveness of our proposed MDTeamGPT across various LLMs, we conducted rigorous experiments to assess its performance. The knowledge bases used in this experiment stored 600 rounds of medical consultation experiences. Figure \ref{fig4} (A) and (B) compare the accuracy improvements of MDTeamGPT versus single-agent baselines for each model. The results indicate that all models benefit from its use, demonstrating both the broad applicability and enhanced effectiveness of the framework. Although the extent of improvement varies with the model’s static knowledge and instruction-following capabilities, our MDTeamGPT framework significantly enhances medical consultation across different LLMs.

\subsection{Generalization of the Knowledge Base}

We conducted cross-dataset experiments to evaluate the knowledge base’s generalization and the framework’s overall effectiveness. Using \textit{CorrectKB} and \textit{ChainKB} (constructed from 600 consultation rounds) with gpt-4-turbo as the backbone, we stored outcomes from one dataset (e.g., MedQA) in the knowledge bases and then tested them on another dataset (e.g., PubMedQA).
\begin{table}[!htbp]
\centering
\renewcommand{\arraystretch}{1.4} % 增加行间距
\setlength{\tabcolsep}{6pt} % 调整列间距
\Large % 调整整体表格字体大小
\resizebox{\columnwidth}{!}{%
\begin{tabular}{c|cc|cc|cc}
\specialrule{2pt}{0pt}{0pt} % 加粗最上方的横线
             & \multicolumn{2}{c|}{Vanilla} & \multicolumn{2}{c|}{\begin{tabular}[c]{@{}c@{}}MedQA\\ (\textit{CorrectKB}, \textit{ChainKB})\end{tabular}} & \multicolumn{2}{c}{\begin{tabular}[c]{@{}c@{}}PubMedQA\\ (\textit{CorrectKB}, \textit{ChainKB})\end{tabular}} \\ \hline
             & \begin{tabular}[c]{@{}c@{}}\textbf{Accuracy}\\ (\%)\end{tabular}  & \begin{tabular}[c]{@{}c@{}}\textbf{F1-score}\\ (\%)\end{tabular} & \begin{tabular}[c]{@{}c@{}}\textbf{Accuracy}\\ (\%)\end{tabular} & \begin{tabular}[c]{@{}c@{}}\textbf{F1-score}\\ (\%)\end{tabular} & \begin{tabular}[c]{@{}c@{}}\textbf{Accuracy}\\ (\%)\end{tabular} & \begin{tabular}[c]{@{}c@{}}\textbf{F1-score}\\ (\%)\end{tabular} \\ \hline
MedQA        & 84.3           & 82.1         & 90.1          & 88.4         & 87.9          & 85.7          \\
PubMedQA     & 77.6           & 75.9         & 79.7          & 76.9         & 83.9          & 82.6          \\ 
\specialrule{2pt}{0pt}{0pt} % 加粗最下方的横线
\end{tabular}%
}
\caption{\label{table3} Cross-dataset generalization result.}
\label{tab:table5}
\end{table}

\vspace{-0.5em}

As shown in Table \ref{table3}, when using the \textit{CorrectKB} and \textit{ChainKB} derived from PubMedQA for testing on MedQA, the accuracy increased by 3.6\% compared to the condition without any knowledge base; conversely, when using the \textit{CorrectKB} and \textit{ChainKB} derived from MedQA for testing on PubMedQA, the accuracy increased by 2.1\% relative to the baseline without any knowledge base. These experimental results indicate that the framework not only successfully transfers specific consultation cases but, more importantly, effectively migrates the reflective reasoning and thought processes—abstract cognitive strategies derived from inference and accumulated experience—to new cases. By preserving reasoning patterns, \textit{CorrectKB} and \textit{ChainKB} not only store consultation examples but also encapsulate transferable reasoning capabilities, allowing our framework to generalize across various scenarios.

\section{Conclusion}

In this paper, we propose a novel MDT multi-agent medical consultation framework, MDTeamGPT, which leverages a residual discussion mode and consensus aggregation to minimize information redundancy and improve discussion efficiency. By leveraging two knowledge bases, the framework dynamically improves overall consultation accuracy and, to some extent, demonstrates excellent generalization ability. Although the overall performance of the framework largely depends on the capabilities of the foundational model used to store and utilize consultation experiences—which may constrain its performance—our approach still excels in current medical consultation scenarios.

\section*{Ethical Considerations}

This study utilizes publicly available academic datasets and accesses LLMs via official APIs, ensuring ethical and responsible research practices. Specifically, the ethical considerations that were taken into account are outlined as follows:

\textbf{Public Datasets}: The medical datasets used in this research are publicly released and intended for academic purposes. No personal patient data was involved or processed.

\textbf{Licensed API Usage}: Our use of LLMs adheres to the API usage policies, maintaining compliance with fair use standards and respecting intellectual property rights.

\textbf{Transparency}: We provide detailed descriptions of our experimental methods to ensure the reproducibility of our results, and we are committed to promoting transparency in scientific practices.

\section*{Limitations}

Our study provides a comprehensive set of experiments to assess the performance of the proposed MDTeamGPT. However, several areas could be further improved and explored in future research.

\textbf{Increasing Agent Complexity}: Currently, the agents in our framework are defined quite simply. We believe that by incorporating more advanced techniques, such as Tree-of-Thought \cite{39}, or integrating external tools and knowledge bases, we could significantly enhance the reasoning capabilities of the multi-agent medical consultation system.

\textbf{Expanding the Dataset}: In our experiments, we have used only two datasets—MedQA and PubMedQA. To strengthen the validity of our findings, future work will involve testing on a broader range of datasets, which will help confirm the robustness of our method across different types of data.

\textbf{Testing in Real-World Scenarios}: Our research has mainly focused on experiments with pre-existing datasets. In future work, we aim to collect more real-world consultation data, which will allow us to assess how well our framework performs in practical, real-life settings and explore its ability to generalize beyond controlled experimental environments.

% \section*{Acknowledgments}

% This document has been adapted
% by Steven Bethard, Ryan Cotterell and Rui Yan
% from the instructions for earlier ACL and NAACL proceedings, including those for
% ACL 2019 by Douwe Kiela and Ivan Vuli\'{c},
% NAACL 2019 by Stephanie Lukin and Alla Roskovskaya,
% ACL 2018 by Shay Cohen, Kevin Gimpel, and Wei Lu,
% NAACL 2018 by Margaret Mitchell and Stephanie Lukin,
% Bib\TeX{} suggestions for (NA)ACL 2017/2018 from Jason Eisner,
% ACL 2017 by Dan Gildea and Min-Yen Kan,
% NAACL 2017 by Margaret Mitchell,
% ACL 2012 by Maggie Li and Michael White,
% ACL 2010 by Jing-Shin Chang and Philipp Koehn,
% ACL 2008 by Johanna D. Moore, Simone Teufel, James Allan, and Sadaoki Furui,
% ACL 2005 by Hwee Tou Ng and Kemal Oflazer,
% ACL 2002 by Eugene Charniak and Dekang Lin,
% and earlier ACL and EACL formats written by several people, including
% John Chen, Henry S. Thompson and Donald Walker.
% Additional elements were taken from the formatting instructions of the \emph{International Joint Conference on Artificial Intelligence} and the \emph{Conference on Computer Vision and Pattern Recognition}.

% % Bibliography entries for the entire Anthology, followed by custom entries
% %\bibliography{anthology,custom}
% % Custom bibliography entries only

\bibliography{custom}

\appendix
\newpage
\onecolumn

\section{Prompt Templates}
\label{Prompt Templates}

Here we provide simplified prompt templates for some of the role Agents to facilitate understanding and demonstration.

\subsection{Primary Care Doctor Agent}
\begin{center}
\scriptsize
\begin{tcolorbox}
[colback=cyan!10!white, colframe=cyan!60!black, title=Primary Care Doctor Agent's Prompt Template, fontupper=\fontsize{10pt}{12pt}\selectfont, fonttitle=\fontsize{10pt}{12pt}\selectfont]

\textbf{Role Description:} \ You are a Primary Care Doctor Agent (triage doctor) responsible for assigning the appropriate doctors to diagnose and treat patients. Each case involves a specific combination of doctors. Based on the symptoms and signs of the patient, choose the most suitable combination of doctors. The selection should include the following roles: General Internal Medicine Doctor, General Surgeon, Pediatrician, Obstetrician and Gynecologist, Radiologist, Neurologist, Pathologist, Pharmacist, with Radiologist, Pathologist, and Pharmacist being mandatory agents, while other agent roles are assigned based on specific patient conditions.

\textbf{Output Format:} \ Before selecting the roles, you are required to provide reasons for your choice: \

Why are these doctors selected based on the patient's symptoms, signs, and history? \
How will each selected doctor contribute to the diagnosis and treatment of the patient? \ When making your selection, consider all relevant information about the patient to ensure that all potential issues are covered. The output should be in the following format: \ \texttt{[\{agent1\}, \{agent2\}, \{agent3\}, \{agent4\}, \{agent5\}]}.

\textbf{Examples:} 

\textbf{Input question:} A 23-year-old pregnant woman at 22 weeks gestation presents with burning upon urination. She states it started 1 day ago and has been worsening despite drinking more water and taking cranberry extract. She otherwise feels well and is followed by a doctor for her pregnancy. Her temperature is 97.7°F (36.5°C), blood pressure is 122/77 mmHg, pulse is 80/min, respirations are 19/min, and oxygen saturation is 98\% on room air. Physical exam is notable for an absence of costovertebral angle tenderness and a gravid uterus. Which of the following is the best treatment for this patient? 

\textbf{Reasoning:} Given that the patient is pregnant, the burning sensation on urination may indicate a urinary tract infection, and an obstetrician should be consulted to rule out pregnancy-specific complications. A radiologist may need to check for other potential issues with the urinary system, while the pathologist and pharmacist can provide laboratory support and medication suggestions. 

\textbf{Output roles:} 

\texttt{[\{Obstetrician and Gynecologist\}, \{Radiologist\}, \{Pathologist\}, \{Pharmacist\}]}

\textbf{Input question:} A 3-month-old baby died suddenly at night while asleep. His mother noticed that he had died only after she awoke in the morning. No cause of death was determined based on the autopsy. Which of the following precautions could have prevented the death of the baby? 

\textbf{Reasoning:} In this case, the pediatrician should assess the baby's history and any potential health risks. The neurologist can help rule out any neurological abnormalities, while the radiologist and pathologist provide necessary assistance, and the pharmacist can assist with preventive measures. 

\textbf{Output roles:} 

\texttt{[\{Pediatrician\}, \{Neurologist\}, \{Radiologist\}, \{Pathologist\}, \{Pharmacist\}]}

\textbf{Input question:} A mother brings her 3-week-old infant to the pediatrician's office because she is concerned about his feeding habits. He was born without complications and has not had any medical problems up until this time. However, for the past 4 days, he has been fussy, is regurgitating all of his feeds, and his vomit is yellow in color. On physical exam, the child's abdomen is minimally distended but no other abnormalities are appreciated. Which of the following embryologic errors could account for this presentation? 

\textbf{Reasoning:} Due to the vomiting and mild abdominal distention, a digestive system issue may be involved. The pediatrician should first evaluate the overall health of the baby. A general surgeon can assess whether there are any gastrointestinal obstructions, while a radiologist should conduct imaging tests, and the pathologist will help with further diagnosis of any pathological changes. The pharmacist can assist in evaluating medication interventions. 

\textbf{Output roles:} 

\texttt{[\{Pediatrician\}, \{General Surgeon\}, \{Radiologist\}, \{Pathologist\}, \{Pharmacist\}]}

\textbf{Tip:} \ For each new patient case, carefully analyze the patient’s condition and select the most appropriate combination of doctors. You need to balance various medical specialties to ensure all potential health issues are addressed and provide adequate reasoning for your doctor selection to ensure all relevant concerns are covered.
\end{tcolorbox}

\end{center}

\subsection{Specialist Doctor Agents}
Here we only showcase the prompt template for the Obstetrician and Gynecologist. Other specialist doctor agents follow the same template, merely altering the medical background section while maintaining the same logical structure.
\begin{center}
\scriptsize
\begin{tcolorbox}
[colback=cyan!10!white, colframe=cyan!60!black, title=Obstetrician and Gynecologist Agent's Prompt Template, breakable, fontupper=\fontsize{10pt}{12pt}\selectfont, fonttitle=\fontsize{10pt}{12pt}\selectfont]

\textbf{Role Description:} \\
You are an Obstetrician and Gynecologist, specializing in women's reproductive health and pregnancy-related medical issues. Your task is to provide a professional diagnosis and treatment recommendation based on the patient's symptoms and medical history, ensuring the chosen treatment is safe and effective for the pregnant woman.

\vspace{0.5em}

\textbf{Task:} \\
First, identify the current consultation round number \{$i$\}. Each round of discussion follows a consistent Chain-of-Thought (CoT) process. During each round, you will both call and store data in the shared message pool, referencing previous rounds of discussion.

\vspace{0.5em}

\textbf{CoT Process:}
\begin{enumerate}
    \item \textbf{Patient Condition Analysis:} \\
    Carefully read the patient’s description of symptoms, combining their signs, clinical examination, and pregnancy status for a comprehensive analysis.
    \item \textbf{Treatment Option Evaluation:} \\
    Based on your professional knowledge, analyze all available treatment options, paying particular attention to drug safety for both the pregnant woman and the fetus.
    \item \textbf{Store in Historical Shared Pool:} \\
    - \textit{Round} \{\(i\)\}: Store the current round number and diagnosis conclusion in the Historical Shared Pool.
    \item \textbf{Call Historical Shared Pool:} \\
    - \textit{Round} \{\(i+1\)\}: Call the Historical Shared Pool from rounds \{\(i-1\)\} and \{\(i-2\)\} to reference the previous two rounds' discussion content for further analysis.
    \item \textbf{Refine or Confirm Based on Feedback:} \\
    Integrate the feedback from the Historical Shared Pool, reassessing your treatment plan. Adjust the selection if necessary; otherwise, confirm the previous choice and explain the rationale.
    \item \textbf{Select Optimal Treatment Plan:} \\
    Determine the most appropriate treatment for the patient and explain your decision.
    \item \textbf{Express Conclusion:} \\
    Use the following format to express your conclusion: \\
    \texttt{Choice: \{Option ID\}: \{Option Content\}} \\
    Example: \texttt{Choice: \{E\}: \{Nitrofurantoin\}}
\end{enumerate}

\vspace{0.5em}

\textbf{MDT Consultation Flow:}

\vspace{0.5em}

\underline{Round \{1\} Discussion:}
\begin{itemize}
    \item \textit{Patient Condition Analysis}
    \item \textit{Treatment Option Evaluation}
    \item Store in Historical Shared Pool
    \item Select Optimal Treatment Plan
    \item Express Conclusion
\end{itemize}

\underline{Round \{2\} Discussion:}
\begin{itemize}
    \item \textit{Patient Condition Analysis}
    \item \textit{Treatment Option Evaluation}
    \item Call Historical Shared Pool from Round \{\(i-1\)\}
    \item Store in Historical Shared Pool
    \item Refine or Confirm Based on Feedback
    \item Select Optimal Treatment Plan
    \item Express Conclusion
\end{itemize}

\underline{Round \(\{i+1\} \geq 3\) Discussion:}
\begin{itemize}
    \item \textit{Patient Condition Analysis}
    \item \textit{Treatment Option Evaluation}
    \item Call Historical Shared Pool from Rounds \{\(i-1\)\} and \{\(i-2\)\}
    \item Store in Historical Shared Pool
    \item Refine or Confirm Based on Feedback
    \item Select Optimal Treatment Plan
    \item Express Conclusion
\end{itemize}

\textbf{Example Prompt:}

\vspace{0.5em}

\textbf{Question:}  A 23-year-old pregnant woman at 22 weeks gestation presents with burning upon urination. She states it started 1 day ago and has been worsening despite drinking more water and taking cranberry extract. She otherwise feels well and is followed by a doctor for her pregnancy. Her temperature is 97.7°F (36.5°C), blood pressure is 122/77 mmHg, pulse is 80/min, respirations are 19/min, and oxygen saturation is 98\% on room air. Physical exam is notable for an absence of costovertebral angle tenderness and a gravid uterus. Which of the following is the best treatment for this patient?

\textbf{Options}: 
\{"A": "Ampicillin", "B": "Ceftriaxone", "C": "Ciprofloxacin", "D": "Doxycycline", "E": "Nitrofurantoin"\}

\vspace{0.5em}

\textit{Round \(\{1\}\) Discussion Example:}
\begin{enumerate}
    \item \textit{Patient Condition Analysis:} The patient is 22 weeks pregnant, presenting symptoms consistent with a urinary tract infection (UTI).
    \item \textit{Treatment Option Evaluation:} Considering pregnancy, medication choice should be cautious. Nitrofurantoin is a relatively safe option for UTI in pregnancy.
    \item Store in Historical Shared Pool: Store diagnosis and round information.
    \item Select Optimal Treatment Plan: Nitrofurantoin offers a balance of safety and efficacy for the patient's condition.
    \item Express Conclusion: \texttt{Choice: \{E\}: \{Nitrofurantoi\}}
\end{enumerate}

\textit{Round \(\{2\}\) Discussion Example:}
\begin{enumerate}
    \item \textit{Patient Condition Analysis:} The patient continues to show stable vital signs, consistent with a UTI.
    \item \textit{Treatment Option Evaluation:} Nitrofurantoin remains a safe and effective choice.
    \item Call Historical Shared Pool: Review feedback from other doctors in Round 1.
    \item Store in Historical Shared Pool: Store the second-round decision and reasoning.
    \item Refine or Confirm Based on Feedback: Feedback from other doctors supports the selection of Nitrofurantoin, so the choice remains unchanged.
    \item Express Conclusion: \texttt{Choice: \{E\}: \{Nitrofurantoin\}}
\end{enumerate}

\textit{Round \(\{i+1\} \geq 3\) Discussion Example:}
\begin{enumerate}
    \item \textit{Patient Condition Analysis:} Same as previous rounds; the symptoms remain consistent with a UTI.
    \item \textit{Treatment Option Evaluation:} Nitrofurantoin is still the most appropriate choice.
    \item Call Historical Shared Pool: Reference prior round discussions for further feedback.
    \item Store in Historical Shared Pool: Store the third-round decision and reasoning.
    \item Refine or Confirm Based on Feedback: After reviewing previous rounds, Nitrofurantoin remains the confirmed choice.
    \item Express Conclusion: \texttt{Choice: \{E\}: \{Nitrofurantoin\}}
\end{enumerate}
\end{tcolorbox}

\end{center}

\subsection{Lead Physician Agent}

\begin{center}
\scriptsize
\begin{tcolorbox}
[colback=cyan!10!white, colframe=cyan!60!black, title=Lead Physician Agent's Prompt Template, breakable, fontupper=\fontsize{10pt}{12pt}\selectfont, fonttitle=\fontsize{10pt}{12pt}\selectfont]

\textbf{Role Description:} \\
You are a Lead Physician responsible for organizing and summarizing the diagnoses and treatment recommendations provided by specialist doctor agents in each consultation round. Your primary task is to classify all responses from specialists and store them in the \textbf{Historical Shared Pool} for future reference. You do not directly participate in diagnosis but ensure that discussions remain structured, logical, and coherent across multiple rounds.

\textbf{Task:} \\
At the end of each consultation round \{\(i\)\}, you must perform the following tasks:

\begin{enumerate}
    \item \textbf{Collect Responses from All Specialists:} \\
    You will receive responses from all specialists in the current consultation round. Each specialist’s response follows this format:
    \begin{verbatim}
    {
      "{Agent_Name}": {
        "Reasoning": "{Explanation of diagnosis and treatment choice}"
        "Choice": "{Option ID}: {Option Content}"
      }
    }
    \end{verbatim}

    \item \textbf{Categorize and Organize Responses:} \\
    Based on the collected responses, systematically classify the information into the following four categories:
    \begin{itemize}
        \item \textbf{Consistency:} Identify diagnoses and treatment recommendations that are similar or identical across multiple specialists.
        \item \textbf{Conflict:} Highlight contradictory diagnoses or treatment choices among specialists and specify the conflicting points.
        \item \textbf{Independence:} Extract unique insights mentioned by individual specialists that are not addressed by others.
        \item \textbf{Integration:} Synthesize a structured summary that incorporates all perspectives, balancing consensus, disagreements, and unique viewpoints.
    \end{itemize}

    \item \textbf{Store Processed Data in Historical Shared Pool:} \\
    Once categorized, save the structured information in the \textbf{Historical Shared Pool} in JSON format:
    \begin{verbatim}
{
…
  "round {i}": {
    "consistency": [
      "{...}"
    ],
    "conflict": [
      "{...}"
    ],
    "independence": [
      "{...}"
    ],
    "integration": [
      "{...}"
    ]
  },
…
}
    \end{verbatim}
\end{enumerate}

\textbf{Process Flow:} \\

\underline{Step 1: Collect Responses from Specialist Doctors} \\
Gather responses from all specialists in the current consultation round:
\begin{verbatim}
{
  "Obstetrician and Gynecologist": {
    "Choice": "E: Nitrofurantoin",
    "Reasoning": "[...]"
  },
  "Pathologist": {
    "Choice": "B: Cephalexin",
    "Reasoning": "[...]"
  }
  ...
}
\end{verbatim}

\underline{Step 2: Categorize Responses} \\
Process all responses into the following categories:

\begin{itemize}
    \item \textbf{Consistency:} Aggregate similar or identical recommendations.
    \item \textbf{Conflict:} Identify and document differing opinions among specialists.
    \item \textbf{Independence:} Extract insights that were mentioned by only one specialist.
    \item \textbf{Integration:} Formulate a structured summary incorporating all perspectives.
\end{itemize}

\underline{Step 3: Store Data in Historical Shared Pool} \\
Once categorized, save the processed data in JSON format:
\begin{verbatim}
{
...
  "round {i}": {
    "consistency": [
      "{Summarize the common aspects found across multiple agents.}"
    ],
    "conflict": [
      "{List conflicting points between agents; leave empty if no conflicts exist.}"
    ],
    "independence": [
      "{Extract unique viewpoints mentioned by only one agent; leave empty if none.}"
    ],
    "integration": [
      "{Provide a well-structured summary integrating all perspectives.}"
    ]
  }
...
}


\end{verbatim}

\end{tcolorbox}

\end{center}

\subsection{Chain-of-Thought Reviewer Agent}
We present the working process of the chain-of-thought reviewer, which has been abstracted and simplified, omitting details that are not suitable for presentation in the paper.

\begin{center}
\scriptsize
\begin{tcolorbox}
[colback=cyan!10!white, colframe=cyan!60!black, title=Chain-of-Thought Reviewer Agent's Prompt Template, breakable, fontupper=\fontsize{10pt}{12pt}\selectfont, fonttitle=\fontsize{10pt}{12pt}\selectfont]

\textbf{Role Description:} \\
You are a Chain-of-Thought Reviewer, responsible for extracting detailed chains of thought from each doctor’s diagnostic process. Your task is to analyze and record the doctors’ thought processes (information from the \textbf{Historical Shared Pool}) and store this information in a vector database for future analysis and retrieval. You will process experience based on whether the consultation outcome is correct or incorrect, storing the relevant information for both correct and incorrect answers.

\textbf{Task:}

\begin{enumerate}
    \item \textbf{Extracting the Chain of Thought:}
    \begin{enumerate}
        \item \textbf{Step 1: Identify the Chain of Thought} \\
        Extract the key reasoning steps from the doctor's discussions and decision-making process, including their initial assumptions, analysis process, and final conclusions.
        
        \item \textbf{Step 2: Structuring the Chain of Thought} \\
        Organize these steps in logical order to form a clear chain of thought, ensuring each step shows how the doctor progressed from one stage to the next.
        
        \item \textbf{Step 3: Recording the Chain of Thought} \\
        Use natural language to document each chain of thought in detail, ensuring it is comprehensive enough for future analysis and reference.
    \end{enumerate}
    
    \item \textbf{Reflecting on Errors:}
    \begin{enumerate}
        \item \textbf{Step 4: Analyze the Mistake} \\
        Identify which steps in the chain of thought might have led to the incorrect diagnosis. Find the root cause, such as faulty assumptions, biased analysis, or overlooked critical information.
        
        \item \textbf{Step 5: Document the Reflection Process} \\
        Clearly indicate the specific reasons for the error and suggest ways to avoid similar mistakes in the future. Highlight which assumptions or analysis steps were wrong and recommend improvements for future reasoning.
    \end{enumerate}
    
    \item \textbf{Storing the Chain of Thought:}
    \begin{enumerate}
        \item \textbf{Step 6: Evaluate the Correctness of the Consultation Outcome} \\
        Determine whether the consultation outcome is correct or incorrect. This process involves retrieving the most similar cases from the Historical Shared Pool based on cosine similarity calculation. Depending on the correctness of the diagnosis, choose the appropriate branch for execution below.
        
        \begin{enumerate}
            \item \textbf{If the Diagnosis is Correct:}
            \begin{enumerate}
                \item \textbf{Step 7: Store the Correct Answer} \\
                If the consultation outcome is correct, store the extracted and recorded chain of thought in the \textit{Correct Answer Knowledge Base (CorrectKB)}. The information should include the medical problem (\( Question \)) along with the structured statements from the final consultation round. Only the final round \{S\_final4\} should be used for the correct answer storage.

                The storage format in JSON should be as follows:
                \begin{verbatim}
                {
                    "Question": [{...}],
                    "Answer": [{...}],
                    "Summary of {S_final4}": [{...}]
                }
                \end{verbatim}

            \end{enumerate}
            
            \item \textbf{If the Diagnosis is Incorrect:}
            \begin{enumerate}
                \item \textbf{Step 8: Store the Incorrect Answer} \\
                If the consultation outcome is incorrect, store the extracted and recorded chain of thought (including the initial hypothesis, analysis process, final conclusion, and error reflection) in the \textit{Chain-of-Thought Knowledge Base (ChainKB)}. For incorrect diagnoses, all rounds from the \textit{Historical Shared Pool} should be used for observation to analyze how the reasoning evolved over time. This will help identify the cause of the error and provide better insights into the overall decision-making process.

                The storage format in JSON should be as follows:
                
                \begin{verbatim}
                {
                    "Question": [{...}],
                    "Correct Answer": [{...}],
                    "Initial Hypothesis": [{...}],
                    "Analysis Process": [{...}],
                    "Final Conclusion": [{...}],
                    "Error Reflection": [{...}]
                    
                }
                \end{verbatim}
                
                This structure will include all rounds of reasoning from the Historical Shared Pool, with each round contributing valuable information for the error reflection and analysis.
            \end{enumerate}
        \end{enumerate}
    \end{enumerate}
\end{enumerate}

\end{tcolorbox}

\end{center}

\subsection{Safety and Ethics Reviewer }

\begin{center}
\scriptsize
\begin{tcolorbox}
[colback=cyan!10!white, colframe=cyan!60!black, title=Safety and Ethics Reviewer's Prompt Template, breakable, fontupper=\fontsize{10pt}{12pt}\selectfont, fonttitle=\fontsize{10pt}{12pt}\selectfont]
\textbf{Role Description:} \\
You are a \textbf{Safety and Ethics Reviewer}, responsible for ensuring that the final diagnosis and treatment recommendation are free from harmful information. Your task is to review the doctors' final opinions, filter out any harmful or unethical suggestions, and output a safe conclusion along with the corresponding answer ID.

\textbf{Task:}
\begin{enumerate}
    \item \textbf{Review Final Opinions:} \\
    Review the doctor's final diagnosis and advice to ensure that there are no unethical or harmful information.
    
    \item \textbf{Filter Harmful Information:} \\
    In the final diagnosis, identify any suggestions that are harmful, unethical or unsafe. This may include unsafe medications, incorrect procedures, or advice that could negatively impact a patient's well-being or safety.
    
    \item \textbf{Refine and Confirm Conclusion:} \\
    Refine the diagnosis after filtering out any harmful information to ensure that it is safe and ethical. Verify the final treatment recommendation to ensure that it is in line with the highest standards of medical ethics.
    
    \item \textbf{Output Conclusion and Answer:} \\
    Provide a clean and accurate conclusion by outputting both the answer ID and recommended treatment option as an output in this format:
    
    \texttt{Answer ID: \{Option ID\}: \{Option Content\}}.
\end{enumerate}

\textbf{Example Prompt:} \\
\textbf{Patient Description:}  A 23-year-old pregnant woman at 22 weeks gestation presents with burning upon urination. She states it started 1 day ago and has been worsening despite drinking more water and taking cranberry extract. She otherwise feels well and is followed by a doctor for her pregnancy. Her temperature is 97.7°F (36.5°C), blood pressure is 122/77 mmHg, pulse is 80/min, respirations are 19/min, and oxygen saturation is 98\% on room air. Physical exam is notable for an absence of costovertebral angle tenderness and a gravid uterus. Which of the following is the best treatment for this patient?

\textbf{Safety Review Example:}
\begin{enumerate}
    \item \textbf{Review Final Opinions:} \\
    The doctors suggested several antibiotics for treating the urinary tract infection, including Ciprofloxacin and Nitrofurantoin.
    
    \item \textbf{Filter Harmful Information:} \\
    Ciprofloxacin is not recommended for pregnant women due to potential harm to the fetus. It was identified as unsafe and filtered out from the final recommendation.
    
    \item \textbf{Refine and Confirm Conclusion:} \\
    After filtering out Ciprofloxacin, Nitrofurantoin was confirmed as the safest option for treating the patient's condition during pregnancy.
    
    \item \textbf{Output Conclusion and Answer:} \\
    \texttt{Answer ID: \{E\}: \{Nitrofurantoin\}}
\end{enumerate}

\end{tcolorbox}
\end{center}

\section{Additional Details}
\label{Additional details}
\subsection{Case Studies of the Lead Physician Agent’s Role and Function}
\label{Lead Physician Agent’s Work}
In this section, we will illustrate how the lead physician organizes and refines the discussion outcomes from the current round of Agents into four categories: Consistency, Conflict, Independence, and Integration. All responses from the roles are based on gpt-4-turbo. We will use Case A and Case B as examples to demonstrate this process.

\begin{center}
\scriptsize
\begin{tcolorbox}
[colback=gray!10!white,colframe=black!75!black, title=Lead Physician Agent (Case A), breakable, fontupper=\fontsize{10pt}{12pt}\selectfont, fonttitle=\fontsize{10pt}{12pt}\selectfont]

\textbf{Question:} A 1-year-old boy presents to the emergency department with weakness and a change in his behavior. His parents state that they first noticed the change in his behavior this morning and it has been getting worse. They noticed the patient was initially weak in his upper body and arms, but now he won’t move his legs with as much strength or vigor as he used to. Physical exam is notable for bilateral ptosis with a sluggish pupillary response, a very weak sucking and gag reflex, and shallow respirations. The patient is currently drooling and his diaper is dry. The parents state he has not had a bowel movement in over 1 day. Which of the following is the pathophysiology of this patient’s condition?

\textbf{Options}: 
\{"A": "Antibodies against postsynaptic nicotinic cholinergic ion channels", "B": "Autoantibodies against the presynaptic voltage-gated calcium channels", "C": "Autoimmune demyelination of peripheral nerves", "D": "Blockade of presynaptic acetylcholine release at the neuromuscular junction", "E": "Lower motor neuron destruction in the anterior horn"\}

\vspace{1em}
\noindent\rule{\linewidth}{0.4pt}

\textbf{Pediatrician:} \\

• \textbf{Reasoning:} As a pediatrician, I focus on common neuro-muscular diseases in children. The symptoms of this 1-year-old boy — weakness in both upper and lower limbs, sluggish pupillary response, weakened swallowing and gag reflexes, as well as drooling and a dry diaper — suggest a possible neuro-muscular junction disorder. Given the combination of these symptoms, the most likely pathophysiology is \textcolor{blue}{blockade of presynaptic acetylcholine release at the neuromuscular junction (D).} This is a manifestation of \textcolor{blue}{myasthenia gravis (MG)}, which typically occurs due to immune responses disrupting normal neuromuscular signal transmission. Other options, such as autoimmune demyelination or lower motor neuron destruction, are less likely given the patient’s clinical presentation. \\
• \textbf{Choice:} \{D\} \{Blockade of presynaptic acetylcholine release at the neuromuscular junction\}

\vspace{0.5em}
\noindent\rule{\linewidth}{0.4pt}

\textbf{Neurologist:} \\

• \textbf{Reasoning:} As a neurologist, I would typically consider neuro-muscular diseases in children. The boy’s symptoms, including symmetric ptosis, sluggish pupillary response, and weakened swallowing reflex, strongly suggest \textcolor{blue}{autoantibodies against the presynaptic voltage-gated calcium channels (B).} This immune-mediated pathology is often seen in diseases like \textcolor{blue}{myasthenia gravis.} While other options could also be linked to neuro-muscular disorders, autoantibodies against presynaptic calcium channels cause disruption of neural signal transmission, which is particularly evident in this patient’s symptoms. \\
• \textbf{Choice:} \{B\} \{Autoantibodies against the presynaptic voltage-gated calcium channels\}

\vspace{0.5em}
\noindent\rule{\linewidth}{0.4pt}

\textbf{Pathologist:} \\

• \textbf{Reasoning:} As a pathologist, my role is to analyze the relationship between symptoms and pathological mechanisms. Based on the clinical presentation, the 1-year-old boy’s symptoms, such as bilateral ptosis, sluggish pupillary response, and difficulty swallowing, suggest a dysfunction at the \textcolor{blue}{neuromuscular junction. Blockade of presynaptic acetylcholine release (D)} best explains these symptoms, especially when the immune system interferes with neuromuscular signal transmission. This pathophysiological mechanism is common in \textcolor{blue}{myasthenia gravis} and often leads to muscle weakness and autonomic dysfunction. \\
• \textbf{Choice:} \{D\} \{Blockade of presynaptic acetylcholine release at the neuromuscular junction\}

\vspace{0.5em}
\noindent\rule{\linewidth}{0.4pt}

\textbf{Pharmacist:} \\

• \textbf{Reasoning:} As a pharmacist, I approach the diagnosis from a pharmacological perspective. The boy’s symptoms, including upper and lower limb weakness, ptosis, and drooling, align with \textcolor{blue}{blockade of presynaptic acetylcholine release (D).} In this case, neural signals cannot effectively transmit to the muscles, leading to motor dysfunction. This pathophysiology is similar to \textcolor{blue}{myasthenia gravis}, which is typically immune-mediated and disrupts neuromuscular junction signaling. Other options, such as autoimmune demyelination or lower motor neuron destruction, do not fully align with these symptoms. \\
• \textbf{Choice:} \{D\} \{Blockade of presynaptic acetylcholine release at the neuromuscular junction\}

\vspace{0.5em}
\noindent\rule{\linewidth}{0.4pt}

\textbf{Radiologist:} \\

• \textbf{Reasoning:} As a radiologist, I do not directly participate in treatment selection, but I consider the patient’s symptoms to hypothesize potential causes. Based on the boy’s symptoms, including ptosis, weakness, and weakened swallowing reflex, the most likely pathophysiology is \textcolor{blue}{blockade of presynaptic acetylcholine release at the neuromuscular junction (D).} This mechanism typically involves the immune system interfering with neuromuscular junction signaling, leading to a failure in signal transmission, which aligns with the patient’s clinical presentation. While other options could also be causes, this mechanism is more commonly observed. \\
• \textbf{Choice:} \{D\} \{Blockade of presynaptic acetylcholine release at the neuromuscular junction\}

\vspace{0.5em}
\noindent\rule{\linewidth}{0.4pt}

\textbf{Lead Physician:}\\

\textbf{Consistency:} \\
In this discussion, most experts chose \textcolor{blue}{blockade of presynaptic acetylcholine release at the neuromuscular junction (D)}, especially pathologists, pharmacists, and pediatricians who supported this choice, as they believe it explains the patient’s symptoms such as bilateral ptosis, weakness, and weakened swallowing reflexes. Several experts pointed out that this pathological mechanism aligns with common neuro-muscular diseases such as \textcolor{blue}{myasthenia gravis.} However, despite this consensus, the neurologist chose \textcolor{blue}{autoantibodies against the presynaptic voltage-gated calcium channels (B)}, suggesting an immune-mediated mechanism, though this diverges from the analysis of the other specialists. Overall, the majority of experts’ choices reflect consistency in their focus on neuro-muscular junction dysfunction.

\textbf{Conflict:} \\
The conflict primarily lies in the neurologist’s choice, as he favors \textcolor{blue}{autoantibodies against the presynaptic voltage-gated calcium channels (B)}, believing this mechanism explains the patient’s autoimmune neuropathy. In contrast, other experts, such as the pathologist and pharmacist, recommended \textcolor{blue}{blockade of presynaptic acetylcholine release (D)}, which they argue better aligns with the symptoms, particularly considering the involvement of the neuro-muscular junction. This divergence highlights the difference in emphasis between the neurologist’s focus on autoimmune disease and the other experts’ focus on neuromuscular junction dysfunction.

\textbf{Independence:} \\
Each expert’s choice reflects their professional background and perspective. Pathologists and pharmacists focus on the dysfunction at the neuromuscular junction, recommending \textcolor{blue}{blockade of presynaptic acetylcholine release (D)}, emphasizing its role in diseases like myasthenia gravis. The neurologist, with a focus on autoimmune disorders, believes that \textcolor{blue}{autoantibodies against the presynaptic voltage-gated calcium channels (B)} better explain the patient’s immune-related pathology. Additionally, the radiologist, while not directly involved in drug selection, supports \textcolor{blue}{blockade of presynaptic acetylcholine release (D)}, reflecting the widespread acceptance of this mechanism.

\textbf{Integration:} \\
In summary, while the neurologist’s choice of \textcolor{blue}{autoantibodies against the presynaptic voltage-gated calcium channels (B)} offers a valid alternative, most experts recommend \textcolor{blue}{blockade of presynaptic acetylcholine release at the neuromuscular junction (D)}, which is more commonly observed in patients with neuro-muscular junction dysfunction. Pathologists, pharmacists, and pediatricians agree that this mechanism is the most likely cause of the patient’s symptoms, and it is consistent with the clinical presentation of myasthenia gravis. While the neurologist offers a different perspective based on autoimmune pathology, \textcolor{blue}{blockade of presynaptic acetylcholine release (D)} remains the most widely supported explanation for this case.

\end{tcolorbox}
\end{center}

\begin{center}
\scriptsize
\begin{tcolorbox}
[colback=gray!10!white,colframe=black!75!black, title=Lead Physician Agent (Case B), breakable, fontupper=\fontsize{10pt}{12pt}\selectfont, fonttitle=\fontsize{10pt}{12pt}\selectfont]

\textbf{Question:} A previously healthy 36-year-old man comes to the physician for a yellow discoloration of his skin and dark-colored urine for 2 weeks. He does not drink any alcohol. Physical examination shows jaundice. Abdominal and neurologic examinations show no abnormalities. Serum studies show increased levels of alanine aminotransferase (ALT) and aspartate aminotransferase (AST). A liver biopsy is performed and a photomicrograph after periodic acid-Schiff-staining is shown. Which of the following is the most likely additional finding in this patient?

\textbf{Options}: 
\{"A": "Tropheryma whipplei-specific RNA on PCR", "B": "Bullous changes of the lung bases on chest CT", "C": "Beading of intra- and extrahepatic bile ducts on ERCP", "D": "Myocardial iron deposition on cardiovascular MRI", "E": "Dark corneal ring on slit-lamp examination"\}

\vspace{1em}
\noindent\rule{\linewidth}{0.4pt}

\textbf{General Internal Medicine Doctor:} \\

• \textbf{Reasoning:} As a general internal medicine doctor, I first consider the patient’s symptoms — \textcolor{blue}{jaundice and dark-colored urine} — which strongly suggest the possibility of \textcolor{blue}{liver disease.} Combined with the patient’s \textcolor{blue}{abnormal liver function (elevated ALT and AST)} and the pathological findings from the \textcolor{blue}{liver biopsy}, there is decisive evidence supporting liver dysfunction. Given these findings, \textcolor{blue}{bullous changes of the lung bases on chest CT (B)} are a \textcolor{blue}{common complication} in such cases, especially since systemic conditions like liver disease can affect the lungs. I believe this aligns perfectly with the clinical presentation, so I choose B. \\
• \textbf{Choice:} \{B\} \{Bullous changes of the lung bases on chest CT\}

\vspace{0.5em}
\noindent\rule{\linewidth}{0.4pt}

\textbf{General Surgeon:} \\

• \textbf{Reasoning:} As a general surgeon, I focus on potential complications that may require surgical intervention. The patient’s \textcolor{blue}{liver dysfunction} (indicated by \textcolor{blue}{jaundice} and \textcolor{blue}{elevated liver enzymes}) suggests chronic liver disease, which is known to lead to pulmonary complications. There is \textcolor{blue}{decisive evidence} in the literature that bullous changes at the lung bases can occur in patients with chronic liver conditions such as cirrhosis. This finding is consistent with the patient’s overall clinical picture. Therefore, I also choose B. \\
• \textbf{Choice:} \{B\} \{Bullous changes of the lung bases on chest CT\}

\vspace{0.5em}
\noindent\rule{\linewidth}{0.4pt}

\textbf{Radiologist:} \\

• \textbf{Reasoning:} As a radiologist, my role is to detect and interpret abnormalities in imaging studies. In this case, the patient’s jaundice and \textcolor{blue}{liver dysfunction} suggest a systemic disease that often has pulmonary manifestations. \textcolor{blue}{Chest CT showing bullous changes at the lung bases is a typical imaging feature} in patients with liver disease. The fact that these changes correlate with the \textcolor{blue}{liver biopsy} findings provides \textcolor{blue}{decisive evidence} for this diagnosis. Thus, I choose B. \\
• \textbf{Choice:} \{B\} \{Bullous changes of the lung bases on chest CT\}

\vspace{0.5em}
\noindent\rule{\linewidth}{0.4pt}

\textbf{Lead Physician:}\\

\textbf{Consistency:} \\
In this discussion, all experts unanimously selected \textcolor{blue}{bullous changes of the lung based on chest CT (B).} They all emphasized that the patient’s \textcolor{blue}{liver disease}—as evidenced by \textcolor{blue}{jaundice, elevated ALT and AST}, and \textcolor{blue}{liver biopsy findings}—can have systemic effects, including \textcolor{blue}{pulmonary complications}. Particularly, the \textcolor{blue}{bullous changes} seen on chest CT are \textcolor{blue}{decisively linked} to chronic liver conditions (e.g., cirrhosis) and represent a common additional finding. This strong concordance among the experts highlights that the imaging abnormality is well supported by the clinical evidence.

\textbf{Conflict:} \\
No conflict exists.
\{...\}

\textbf{Independence:} \\
No independence exists.
\{...\}

\textbf{Integration:} \\
In summary, all experts agree that \textcolor{blue}{bullous changes of the lung bases on chest CT (B)} is the most likely additional finding. The decisive evidence—including \textcolor{blue}{jaundice, elevated liver enzymes}, and \textcolor{blue}{liver biopsy findings}—strongly supports the presence of liver disease, which is known to cause pulmonary complications such as \textcolor{blue}{bullous changes at the lung bases.} This imaging finding, characteristic of chronic liver disease (for instance, cirrhosis), reinforces the overall diagnosis and is consistently endorsed by all experts, thereby confirming its relevance in the clinical scenario.

\end{tcolorbox}
\end{center}

\subsection{Specialist Selection Distribution}
\label{Specialist Selection Distribution}

\begin{figure*}[!htbp]
\centerline{\includegraphics[width=0.9\textwidth]{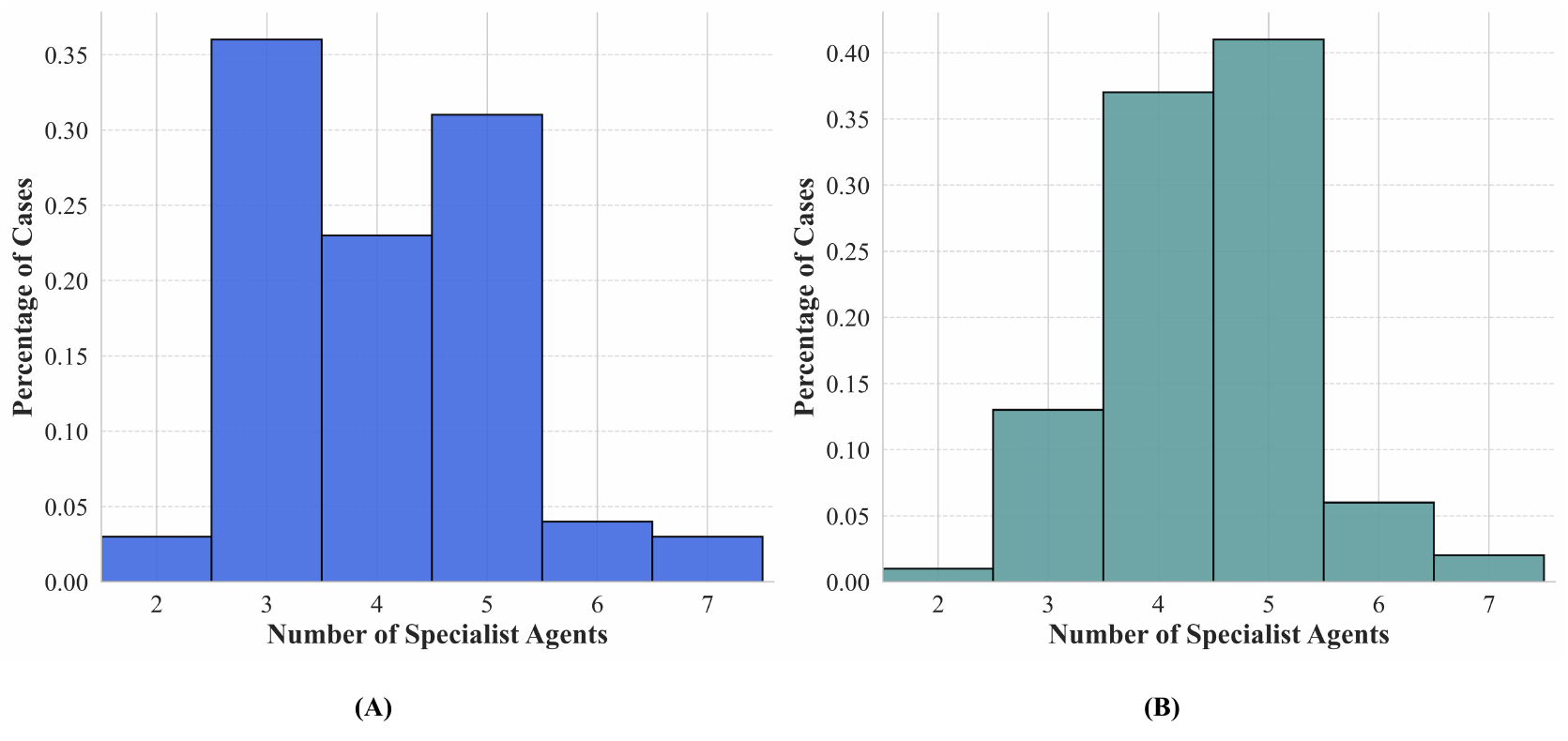}}
  \captionsetup{skip=0pt} % 设置标题与正文的间距为0
  \caption {Distribution of Specialist Selection by the Primary Care Doctor Agent:(A) Distribution on MedQA dataset. (B) Distribution on PubMedQA dataset.}
\label{figapp}
\end{figure*}

This section presents the distribution of specialists arranged by the Primary Care Doctor Agent in our experiments on the MedQA and PubMedQA datasets. The framework includes seven specialist roles: General Internal Medicine Doctor, General Surgeon, Pediatrician, Obstetrician and Gynecologist, Radiologist, Neurologist, Pathologist, and Pharmacist. Figure \ref{figapp} shows the distribution of selected specialists. In the MedQA dataset, the Primary Care Doctor Agent typically assigns around 3 to 5 specialists, whereas in PubMedQA, the selection is more centered around 5. Cases with only 2 or as many as 7 specialists are uncommon, and no instances involve selecting just one specialist.

\subsection{Human Evaluation}
\label{Human evaluation}

\begin{table*}[!htbp]
\centering

\resizebox{\textwidth}{!}{\begin{tabular}{l|l|l}
\hline
\textbf{Evaluation Criterion} & \textbf{Description}                                                                                                                                                                                                                                                                                                                                                                                                                           & \textbf{\begin{tabular}[c]{@{}l@{}}Score \\ (0-4)\end{tabular}} \\ \hline
\textbf{Accuracy}             & \begin{tabular}[c]{@{}l@{}}The correctness of the system’s diagnosis or suggestion.\\ 0 = Completely wrong \\ 1 = Partially correct with major mistakes \\ 2 = Partially correct with minor mistakes \\ 3 = Mostly correct with small gaps \\ 4 = Fully correct\end{tabular}                                                                                                                                                                   &                                                                 \\ \hline
\textbf{Explainability}       & \begin{tabular}[c]{@{}l@{}}Evaluate how clearly the system explains its reasoning. Does the explanation make sense? \\ 0 = No explanation or completely unclear \\ 1 = Explanation lacks logic or clarity \\ 2 = Explanation is mostly clear but somewhat confusing \\ 3 = Clear explanation with good logic but some details could be refined \\ 4 = Clear, detailed, and logical explanation\end{tabular}                                    &                                                                 \\ \hline
\textbf{Completeness}         & \begin{tabular}[c]{@{}l@{}}Rate the system’s response in terms of how comprehensive it is. Does the system consider all relevant factors and possible diagnoses? \\ 0 = Lacks important information and key factors \\ 1 = Missing some key information \\ 2 = Covers some key factors but leaves out a few \\ 3 = Covers most of the relevant factors with minor omissions \\ 4 = Comprehensive and covers all important factors\end{tabular} &                                                                 \\ \hline
\textbf{Rationality}          & \begin{tabular}[c]{@{}l@{}}Evaluate if the system’s suggestion is logical and grounded in medical knowledge. \\ 0 = Completely illogical \\ 1 = Partially illogical with some reason \\ 2 = Mostly logical but with some issues \\ 3 = Reasonable and generally follows medical knowledge \\ 4 = Fully logical and based on solid medical knowledge\end{tabular}                                                                               &                                                                 \\ \hline
\textbf{Diversity}            & \begin{tabular}[c]{@{}l@{}}Does the system provide multiple potential diagnostic paths or solutions? \\ 0 = No diversity, always gives the same answer \\ 1 = Minimal diversity \\ 2 = Few variations, but answers are similar \\ 3 = Provides multiple distinct solutions or perspectives \\ 4 = Offers diverse diagnostic approaches with different viewpoints\end{tabular}                                                                  &                                                                 \\ \hline
\textbf{Efficiency}           & \begin{tabular}[c]{@{}l@{}}Rate the time efficiency of the system. How quickly does it generate responses, especially during multi-round discussions? \\ 0 = Extremely slow, cannot provide useful results in reasonable time \\ 1 = Slow, long wait times \\ 2 = Acceptable, but sometimes a bit slow \\ 3 = Usually efficient and provides results quickly \\ 4 = Fast and provides valid answers almost immediately\end{tabular}            &                                                                 \\ \hline
\textbf{Consistency}          & \begin{tabular}[c]{@{}l@{}}Evaluate whether the system's responses are consistent across different rounds of consultation. \\ 0 = Completely inconsistent \\ 1 = Mostly inconsistent, with some contradictions \\ 2 = Some inconsistencies, but mostly similar answers \\ 3 = No contradictions, consistent throughout \\ 4 = Completely consistent with no contradictions\end{tabular}                                                        &                                                                 \\ \hline
\textbf{User Experience}      & \begin{tabular}[c]{@{}l@{}}Rate your overall experience interacting with the system. Is it easy to use and intuitive? \\ 0 = Difficult to use and navigate \\ 1 = Poor user experience \\ 2 = Fairly intuitive but some issues \\ 3 = Easy to use and navigate, good experience \\ 4 = Extremely smooth, intuitive, and user-friendly\end{tabular}                                                                                             &                                                                 \\ \hline
\textbf{Safety}               & \begin{tabular}[c]{@{}l@{}}Rate the safety of the system's suggestions. Did it ever provide potentially unsafe or harmful advice? \\ 0 = Provided dangerous or unsafe advice \\ 1 = Some potential risks that need attention \\ 2 = Slightly unsafe, requires confirmation \\ 3 = Generally safe but needs further confirmation \\ 4 = Completely safe, no risks or errors\end{tabular}                                                        &                                                                 \\ \hline
\textbf{Innovation}           & \begin{tabular}[c]{@{}l@{}}Rate the level of innovation in the system’s diagnostic approach. Did it offer creative or new ideas? \\ 0 = No innovation, completely conventional \\ 1 = Some new ideas but mostly conventional \\ 2 = Some innovative solutions \\ 3 = Significant innovation and new ideas \\ 4 = Very creative and novel approach with new ways of thinking\end{tabular}                                                       &                                                                 \\ \hline

\end{tabular}}

\caption{\label{tabel4}  MDTeamGPT Human Evaluation Questionnaire.}
\end{table*}

To comprehensively evaluate the performance of MDTeamGPT, we invited five volunteers to participate in the tests. All volunteers hold at least a bachelor’s degree in clinical medicine, and two of them have medical practitioner qualifications. We randomly selected five questions from the MedQA dataset and five questions from the PubMedQA dataset, resulting in a total of 10 test questions. Volunteers observed the outputs of MDTeamGPT and rated them using the survey questionnaire shown in Table~\ref{tabel4}. Each volunteer was assigned integer scores for every test. The evaluation covered 10 dimensions: accuracy, explainability, completeness, rationality, diversity, efficiency, consistency, user experience, safety, and innovation. This portion of the testing was conducted using gpt4-turbo.

As shown in Figure~\ref{figapp}, the volunteers assigned relatively low scores to the "Efficiency" metric, primarily due to the increased time required for multi-round discussions among multiple roles compared to the Single-Agent approach. The "Diversity" metric also received an average score below 3. This may be attributed to the Temperature setting of 0.7, which was chosen to balance determinism and randomness; increasing this value could potentially enhance diversity. For "Completeness", the average score was 3, though some feedback indicated that certain responses lacked sufficient consideration. To address this, future improvements will focus on designing more comprehensive prompts and optimizing the framework’s efficiency. All other metrics received scores above 3, demonstrating the potential value of MDTeamGPT in practical medical applications.

\begin{figure*}[t]
\centerline{\includegraphics[width=0.9\textwidth]{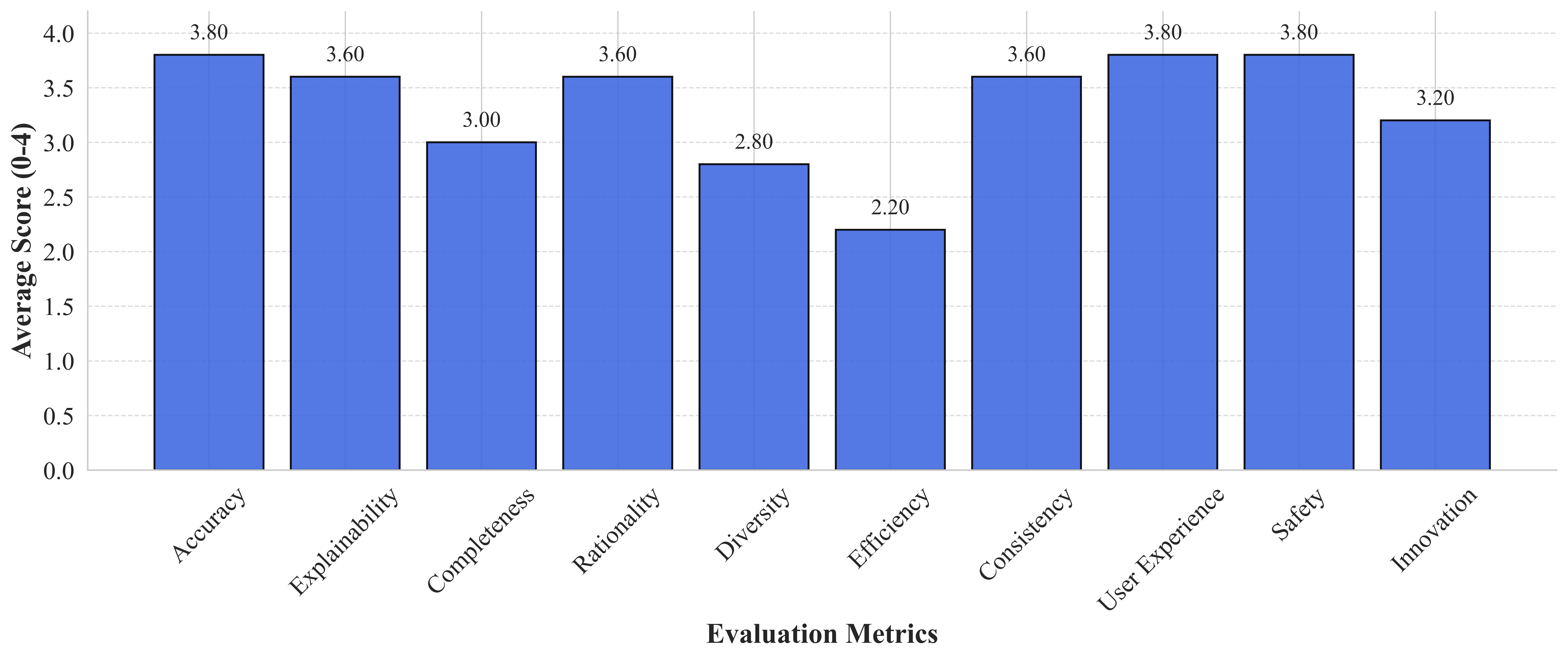}}
  \captionsetup{skip=0pt} % 设置标题与正文的间距为0
  \caption {Human Evaluation: Average Scores of MDTeamGPT.}
\label{figapp}
\end{figure*}

\end{document}